\documentclass[11pt]{article}

\usepackage[final]{acl}
\usepackage{amsmath}
\usepackage{times}
\usepackage{latexsym}
\usepackage{booktabs}
\usepackage{array}
\usepackage{tabularx}
\usepackage{CJKutf8}
\usepackage{multirow}
\usepackage{pifont}
\usepackage[T1]{fontenc}
\usepackage[utf8]{inputenc}

\usepackage{microtype}

\usepackage{inconsolata}
\usepackage{caption}

\usepackage{graphicx}

\title{G-IdiomAlign: A Gloss-Pivoted Benchmark for \\Cross-Lingual Idiom Alignment}

\author{
       Fengying Ye$^{1,}$\footnotemark[1] ~~
       Yanming Sun$^{1,}$\thanks{Equal Contribution.} ~~
       Runzhe Zhan$^1$ ~~ \\
       \textbf{Zheqi Zhang$^2$} ~~
       \textbf{Lidia S. Chao$^1$} ~~
       \textbf{Derek F. Wong}$^{1,}$\thanks{Corresponding Author.}\\
    $^1$NLP$^2$CT Lab, Department of Computer and Information Science, University of Macau \\
    $^2$Faculty of Arts and Humanities, University of Macau \\
    nlp2ct.\{fengying, yanming, runzhe\}@gmail.com, \\ \{lidiasc, derekfw\}@um.edu.mo
    }
\begin{document}
\maketitle

\begin{abstract}
Idioms are difficult to transfer across languages due to their non-compositionality and weak surface-form grounding, making literal mappings unreliable. We present \textbf{G-IdiomAlign}, a gloss-pivoted benchmark where each idiom is anchored by an English gloss from Wiktionary. We further construct a high-confidence reference alignment set for reproducible evaluation. G-IdiomAlign supports two protocols: (1) a controlled {Multiple-Choice Idiom Equivalence} with typed distractors for error attribution; and (2) a {Gloss-Contrastive Generation} contrasting \textit{No-gloss} and \textit{With-gloss} inputs to isolate the effect of an explicit semantic pivot. Across diverse LLMs, a bias to literal translation is a dominant failure mode, especially when the target is a low-resource language. Glosses consistently improve {Gloss-Contrastive Generation} under an embedding-based semantic proxy, but performance remains modest, indicating substantial headroom in the open output space. Subsequent analysis on Qwen3-8B further suggests that cross-condition differences are concentrated more in attention heads than in layers, while better \textit{With-gloss} generations coincide with stronger gloss anchoring\footnote{Dataset:~\url{https://github.com/NLP2CT/G-IdiomAlign}}.
\end{abstract}

\section{Introduction}
Idioms pose a persistent challenge for cross-lingual meaning transfer because their figurative meanings are non-compositional and culturally grounded, making word-by-word composition unreliable \cite{he-etal-2025-investigating}. Recent evidence suggests that large language models (LLMs) often over-index on surface statistical cues (such as collocational frequency or sentence probability) rather than recovering figurative intent from context \cite{mi-etal-2025-rolling, Yang2025EvaluatingLO,Ye-etal-2026-probing}. Accurate cross-lingual idiom alignment is crucial not only for cross-cultural communication and machine-assisted localization, but also as a litmus test for whether LLMs genuinely grasp culturally embedded semantics beyond surface patterns, motivating evaluation protocols that target idiom-to-idiom semantic equivalence over mere lexical overlap.

%However, existing resources provide limited support for controlled evaluation of cross-lingual idiom-to-idiom equivalence. Even strong systems still produce literal, partial, or missing idiom renderings \cite{yang2025evaluating}, while current datasets rarely offer unified benchmarks with standardized protocols for diagnosis. To make meaning explicit and comparable across languages, we adopt English glosses as a shared \emph{semantic pivot}, reflecting an assisted but practical setting in which models can access external meaning support (e.g., lexicons or knowledge bases). We then contrast \textit{No-gloss} and \textit{With-gloss} inputs to isolate the effect of an explicit semantic signal.

However, existing resources offer limited support for controlled and diagnostic evaluation of cross-lingual idiom-to-idiom equivalence. Strong systems frequently produce literal, partial, or missing idiom renderings \cite{Yang2025EvaluatingLO}, yet current datasets lack unified benchmarks with standardized protocols for systematic error attribution. To enable explicit and comparable semantic grounding across languages, we adopt English glosses as a shared \emph{semantic pivot}, modeling a resource-augmented setting where models can leverage external semantic support (e.g., lexicons or knowledge bases). We introduce a contrastive setup between \textit{No-gloss} and \textit{With-gloss} inputs to isolate the effect of this explicit semantic signal.

\newcolumntype{C}{>{\centering\arraybackslash}m{0.20\linewidth}}
\newcolumntype{Y}{>{\centering\arraybackslash}m{0.20\linewidth}}
\newcolumntype{L}{>{\raggedright\arraybackslash}m{0.26\linewidth}}
\newcolumntype{R}{>{\raggedright\arraybackslash}m{0.27\linewidth}}
\begin{CJK}{UTF8}{bsmi}
\begin{table}[t]
\centering
\small
\setlength{\tabcolsep}{3pt}
\begin{tabularx}{\linewidth}{C Y L L}
\toprule
\normalsize\textbf{$I_{\text{source}}$} &
\normalsize\textbf{$I_{\text{target}}$} &
\multicolumn{1}{c}{\normalsize\textbf{$G_{\text{source}}$}} &
\multicolumn{1}{c}{\normalsize\textbf{$G_{\text{target}}$}} \\
\midrule
一鋪清袋 & lose one's shirt
& lose all money at once
& lose all of the money \\
\cmidrule(r){1-4}
mouton de Panurge & follow ... off a cliff
& blindly follow others
& follow a leader blindly \\
\cmidrule(r){1-4}
守口如瓶 & tenir sa langue
& keep one's mouth shut
& hold one's tongue \\
\bottomrule
\end{tabularx}
\caption{Example idiom pairs from \textbf{G-IdiomAlign} across different language pairs. $I$ and $G$ denote idioms and their English glosses; subscripts indicate languages.}
\label{tab: example_idioms}
\end{table}

In this work, we introduce \textbf{G-IdiomAlign}, a gloss-pivoted idiom alignment benchmark across nine core languages, with coverage of four languages from underrepresented families, where each idiom is linked to a meaning-equivalent English gloss from Wiktionary~(examples are shown in Table~\ref{tab: example_idioms}). We construct a high-confidence~reference set via a precision-first pipeline that combines distribution-aware filtering with bidirectional one-to-one constraints. On top of this benchmark, we provide two complementary evaluation settings: a Multiple-Choice Idiom Equivalence task with typed distractors, and a Gloss-Contrastive Generation under \textit{No-gloss} and \textit{With-gloss} inputs.

Across diverse LLMs, models show a pervasive bias to literal translations. Although adding glosses yields consistent but limited improvements under an embedding-based semantic proxy, this underscores the difficulty of producing canonical, meaning-equivalent idioms in an unconstrained space. Subsequent attention-based correlational analyses on Qwen3-8B further suggest that improved \textit{With-gloss} generations align with stronger gloss anchoring, with cross-condition differences concentrating mainly at the level of attention heads rather than broad layer-level shifts. Our analysis positions G-IdiomAlign as a foundation for future work on robust cross-lingual idiom modeling.

Our contributions are as follows: (1) We release \textbf{G-IdiomAlign}, a gloss-pivoted dataset covering 36 language pairs across nine core languages (18{,}785 idiom pairs), filtered via a bidirectional pipeline to ensure high semantic equivalence. (2) We establish two diagnostic protocols for cross-lingual idiom alignment: {Multiple-Choice Idiom Equivalence} with typed distractors, and {Gloss-Contrastive Generation} that contrasts \textit{No-gloss} and \textit{With-gloss} inputs to test semantic grounding. (3) We reveal a widespread bias towards literal translation in LLMs and provide attention-based evidence linking gloss usage to improved semantic anchoring.

\section{Related Work}
\subsection{Idiom Benchmarks}
Existing idiom benchmarks support two core tasks: detection, which identifies whether a phrase is used idiomatically, and disambiguation, which resolves whether an expression should be interpreted literally or figuratively. 
Detection datasets test whether a model distinguishes idiomatic from literal usages, ranging from linguist-curated contrastive sets \cite{mi-etal-2025-rolling} to English test suites such as IdioTS \cite{de-luca-fornaciari-etal-2024-hard}, multilingual benchmarks ID10M \cite{tedeschi-etal-2022-id10m} and CLCL framework \cite{zhou-etal-2023-clcl}. 
Disambiguation datasets including EPIE \cite{saxena2020epie}, MAGPIE \cite{haagsma-etal-2020-magpie}, and MultiCoPIE \cite{sentsova-etal-2025-multicopie} label potential idiomatic expressions with literal versus idiomatic readings, supporting contextual sense selection \cite{fakharian-cook-2021-contextualized,zhou-etal-2021-pie}. Complementary resources broaden coverage further: LIdioms \cite{moussallem-etal-2018-lidioms} links idioms across languages as linked data, and \citet{fu-etal-2025-chengyu} evaluate Chinese idioms across multiple competencies. 
While these efforts are valuable for idiom recognition and interpretation, they do not directly target \emph{idiom-to-idiom} meaning-equivalence alignment within a unified cross-lingual evaluation.

\subsection{Idiom Alignment}
Cross-lingual idiom alignment remains challenging because figurative meanings often diverge from literal forms and are shaped by language- and culture-specific conventions \cite{moussallem-etal-2018-lidioms,donthi-etal-2025-improving}. Recent evaluations confirm persistent failures in both NMT systems and LLMs \cite{Yang2025EvaluatingLO,sun2026exposing}, prompting approaches that decompose translation into semantic analysis and candidate selection \cite{qian-2024-automating} or inject external signals, such as retrieval-augmented MT with loss weighting \cite{liu-etal-2023-crossing} or multilingual idiom knowledge bases \cite{Li_Chen_Yuan_Wu_Yang_Tao_Xiao_2024}. However, these methods often rely on surface-level cues: \citet{sentsova-etal-2025-multicopie} report substantially higher performance on idioms with direct English lexical counterparts, and cross-lingual evaluations note strong prompt sensitivity and performance gaps in low-overlap language pairs \cite{khoshtab-etal-2025-comparative}. This reliance is exacerbated in retrieval-based alignment frameworks like bilingual lexicon induction (BLI), which formulate cross-lingual matching as nearest-neighbor search over candidate sets \cite{li-etal-2023-bilingual}. Such approaches are prone to false positives \cite{Ding_Cao_Zhao_2024}, unless constrained by precision-oriented criteria like bidirectional agreement. To address these limitations, we move beyond surface-driven retrieval by anchoring alignment in meaning-equivalent English glosses and adopt bidirectional constraints to ensure high-precision idiom-to-idiom pairing, thus enabling controlled evaluation that isolates semantic equivalence from lexical shortcuts.

%\section{G-IdiomAlign}
%Idioms are highly non-compositional and culturally grounded, making cross-lingual alignment particularly challenging: surface-form overlap and literal translation are often insufficient to capture semantic equivalence across languages. To address this challenge, we introduce \textbf{G-IdiomAlign}, a gloss-pivoted benchmark for cross-lingual idiom alignment. G-IdiomAlign covers nine languages and leverages the English glosses provided in Wiktionary as a shared semantic pivot to align idioms across languages. We construct the benchmark through a high-precision pipeline that crawls idiom entries and their glosses, generates cross-lingual alignment candidates via gloss embeddings, and confirms gold alignments using distribution-aware filtering and mutual nearest neighbor constraints. G-IdiomAlign is designed for evaluation and diagnostic analysis, and serves as a unified gold reference for both the multiple-choice and open-ended generation settings in our experiments.

\begin{figure*}[t]
  \centering
  \includegraphics[width=0.9\linewidth]{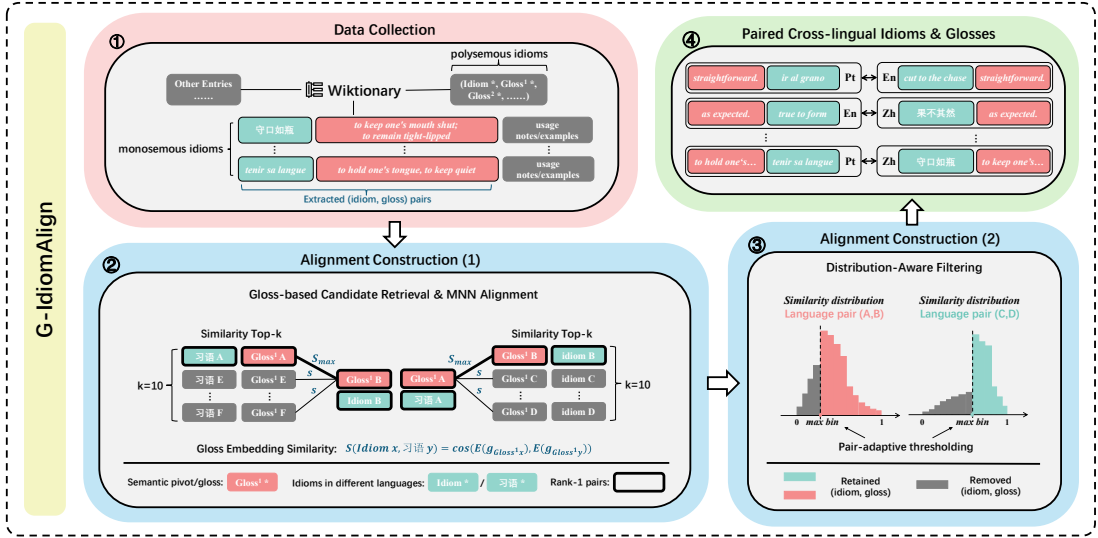}
\caption{Overview of the G-IdiomAlign construction pipeline. Using English glosses as a shared semantic pivot, we extract idiom entries and core glosses from Wiktionary, retrieve top-$k$ candidates in a gloss-embedding space, keep MNN pairs, and apply a pair-specific distribution-aware filter, yielding the final G-IdiomAlign benchmark.}
  \label{fig:pipeline_overview}
\end{figure*}

\section{G-IdiomAlign}
We introduce \textbf{G-IdiomAlign}, a gloss-pivoted benchmark for cross-lingual idiom alignment across nine core languages, with coverage of four languages from low-resource language families.
English glosses from Wiktionary\footnote{https://www.wiktionary.org/} serve as a shared semantic pivot, supporting cross-lingual meaning comparison while mitigating shortcuts based on surface lexical overlap. We construct G-IdiomAlign with a precision-first, staged construction pipeline: (i) extract idiom entries and core glosses (excluding usage notes/examples), (ii) retrieve top-$k$ candidates in a shared gloss-embedding space, (iii) retain mutual nearest neighbor (MNN) pairs via bidirectional agreement, and (iv) apply distribution-aware filtering to produce a high-confidence alignment set. The resulting benchmark is intended for evaluation and diagnostic analysis. Figure~\ref{fig:pipeline_overview} summarizes the construction pipeline.

%\subsection{Language Coverage and Data Collection}
%\textbf{Language Scope.} G-IdiomAlign covers a diverse set of nine languages: Chinese (ZH), English (EN), Finnish (FI), Japanese (JA), Polish (PL), French (FR), German (DE), Spanish (ES), and Portuguese (PT).
%\textbf{Collection Pipeline.} We source data from Wiktionary by traversing category pages explicitly labeled with the \textit{Idiom} tag. To ensure high-quality semantic representations, we implement a targeted parsing strategy that extracts the core definition (English gloss) while discarding noise such as usage notes and examples. These glosses serve as the semantic pivot for our subsequent alignment.
%\textbf{Single-Sense Filtering.} A critical step in our pipeline is the application of \textbf{single-sense filtering}. We strictly retain idioms associated with a unique English gloss, excluding polysemous entries to avoid one-to-many ambiguity. While this reduces the total volume, it prioritizes precision over coverage, establishing a rigorously clean inventory essential for a reliable evaluation benchmark.

\subsection{Language Coverage \& Data Collection} \label{sec:data_construction}

\paragraph{Language Scope.}
G-IdiomAlign covers nine core languages:  De, En, Es, Fi, Fr, Ja, Pl, Zh, and Pt, for which our extraction pipeline yields sufficient high-quality aligned pairs. To broaden language coverage, we further include four languages (Arabic, Korean, Thai, and Vietnamese) and report their results separately in Appendix~\ref{app:data_stats}.

\paragraph{Collection Pipeline.}
We collect idiom entries from language-specific Wiktionary category pages and extract a cleaned core gloss from each entry’s sense definition, excluding auxiliary material such as usage notes and examples; see Appendix~\ref{app:scrape} for implementation details. These glosses provide a consistent meaning description and function as the semantic pivot throughout construction.

\paragraph{Single-Sense Filtering.}
To preserve interpretability of the reference, we keep idioms with a single Wiktionary sense (one gloss) and remove polysemous entries. This avoids one-to-many sense correspondences that would make idiom-to-idiom equivalence ambiguous at construction time. The resulting reference set is smaller but cleaner, supporting more controlled evaluation and diagnosis.

%For each language pair $A \rightarrow B$, we encode the English glosses using the OpenAI text-embedding-3-large model. For an idiom $x$ with gloss $g_x$ and a target idiom $y$ with gloss $g_y$, we define a gloss similarity score as
%\[s(x,y)=\cos\!\big(E(g_x),\,E(g_y)\big),\] where $E(\cdot)$ is the embedding model. We retrieve the top-$k$ candidates for each source idiom according to $s(x,y)$ (we set $k=10$), yielding an initial candidate pool based on semantic proximity.
\begin{table*}[t]
\centering
\small
\setlength{\tabcolsep}{8pt}
\begin{tabular}{l r r l r r l r r l r r}
\toprule
\textbf{Pair} & \textbf{N} & \textbf{\%} &
\textbf{Pair} & \textbf{N} & \textbf{\%} &
\textbf{Pair} & \textbf{N} & \textbf{\%} &
\textbf{Pair} & \textbf{N} & \textbf{\%} \\
\midrule
De--En & 520  & 2.77 & En--Fi & 693  & 3.69 & Es--Pl & 888  & 4.73 & Fr--Pl & 329  & 1.75 \\
De--Es & 456  & 2.43 & En--Fr & 376  & 2.00 & Es--Pt & 429  & 2.28 & Fr--Pt & 209  & 1.11 \\
De--Fi & 353  & 1.88 & En--Ja & 336  & 1.79 & Es--Zh & 1028 & 5.47 & Fr--Zh & 343  & 1.83 \\
De--Fr & 238  & 1.27 & En--Pl & 1182 & 6.29 & Fi--Fr & 267  & 1.42 & Ja--Pl & 345  & 1.84 \\
De--Ja & 229  & 1.22 & En--Pt & 534  & 2.84 & Fi--Ja & 251  & 1.34 & Ja--Pt & 206  & 1.10 \\
De--Pl & 447  & 2.38 & En--Zh & 1782 & 9.49 & Fi--Pl & 589  & 3.14 & Ja--Zh & 416  & 2.21 \\
De--Pt & 280  & 1.49 & Es--Fi & 569  & 3.03 & Fi--Pt & 333  & 1.77 & Pl--Pt & 432  & 2.30 \\
De--Zh & 509  & 2.71 & Es--Fr & 336  & 1.79 & Fi--Zh & 655  & 3.49 & Pl--Zh & 1156 & 6.16 \\
En--Es & 1114 & 5.93 & Es--Ja & 306  & 1.63 & Fr--Ja & 186  & 0.99 & Pt--Zh & 463  & 2.46 \\
\bottomrule
\end{tabular}
\caption{G-IdiomAlign language-pair composition. \textbf{N} denotes the count of aligned idiom pairs and \textbf{\%} denotes the proportion of the dataset (out of all aligned pairs). We report each pair once using a canonical ordering.}
\label{tab:pair_stats}
\end{table*}

\paragraph{Gloss-based Candidate Retrieval.}
For each directed language pair $A \rightarrow B$, we embed the glosses associated with idioms in both languages using the OpenAI text-embedding-3-large \cite{openai2024embedding3}. For a source idiom $x$ with gloss $g_x$ and a candidate idiom $y$ with gloss $g_y$, we define gloss similarity as
\[
s(x,y)=\cos\!\big(E(g_x), E(g_y)\big),
\]
where $E(\cdot)$ denotes the embedding function.
For each $x$, we retrieve the top-$k$ candidates in $B$ by $s(x,y)$ with $k=10$, producing a candidate set for subsequent bidirectional filtering.
Although final alignments are determined by rank-1 agreement (see below), using $k>1$ improves candidate coverage and robustness to embedding noise before enforcing one-to-one constraints.

\paragraph{MNN Alignment.}
To obtain unambiguous evaluation pairs, we enforce a one-to-one matching constraint via mutual nearest neighbors (MNN). We retain a pair $(x,y)$ if and only if $x$ and $y$ are rank-1 nearest neighbors of each other under both directions ($A \rightarrow B$ and $B \rightarrow A$).
This bidirectional criterion removes asymmetric or many-to-one associations that may arise from retrieval artifacts, ensuring that retained alignments reflect strong mutual semantic correspondence.

\paragraph{Distribution-Aware Filtering.}
Since similarity score scales differ substantially across language pairs, fixed global thresholds can be poorly calibrated. Moreover, even under MNN, nearest-neighbor retrieval always returns a best match within the dataset, which can force alignments even when no true equivalent exists, leading to spurious pairs. For example, a Chinese idiom ``洞房花燭夜'' (gloss: the wedding night) may be aligned with the English idiom ``white marriage'' (gloss: an unconsummated marriage): although their glosses share salient words (wedding and marriage), the underlying meanings are not equivalent. Accordingly, we apply a language-pair-specific, parameter-light cutoff to remove weak matches while preserving high-confidence alignments.

For each language pair, we collect the rank-1 similarity scores of MNN-confirmed pairs and discretize similarity scores within-pair range into 10 equal-width bins. Let $b$ denote the modal bin. We retain pairs whose scores fall in bin $b$ or higher, using the lower edge of the modal bin as a cutoff. Similarity scores are used here as a diagnostic signal for relative strength within each language pair, rather than as an absolute criterion of semantic correctness (details are shown in Appendix~\ref{app:dist_filter}).

To ensure deterministic reporting, we compute similarity scores using a fixed canonical direction for each unordered language pair, while the MNN criterion itself is always enforced bidirectionally.

\subsection{Benchmark Statistics}
\label{sec:benchmark_stats}
G-IdiomAlign comprises 18{,}785 aligned idiom pairs across 36 \emph{unordered} language pairs drawn from nine languages. Although alignments are reported without direction, each pair supports evaluation in either direction (e.g., Zh$\rightarrow$En or En$\rightarrow$Zh). For reporting and aggregation, each unordered language pair is listed once using a canonical ordering. Pair sizes range from 186 to 1{,}782, with a median of 422 (interquartile range: 323--574); the full breakdown is provided in Table~\ref{tab:pair_stats}.

\subsection{Alignment Quality Evaluation}
\label{sec:alignment_quality}

We assess the semantic alignment quality of G-IdiomAlign using both human evaluation and LLM-based majority voting. Sampled pairs are rated on a 3-point scale: 2 denotes equivalent meaning and interchangeability across contexts; 1 denotes partial equivalence, where meanings are close but differ in tone, intensity, or pragmatics; and 0 denotes non-equivalence, where lexical or topical relatedness does not imply semantic equivalence.

For Zh--En idiom pairs, we randomly sample 200 pairs and evaluate them with native speakers and senior Ph.D. students with expertise in relevant languages. For the remaining language pairs, we sample 50 pairs per language pair and score them independently with GPT-5.1 \cite{openai2026gpt51}, Gemini-2.5-Pro \cite{geminiteam2025gemini25}, and Claude-4.5-Haiku \cite{anthropic2025claudehaiku45}. LLM judges are prompted to follow the same annotation instructions as human annotators. We use majority voting as the final label; when all judges disagree, we assign score~1 to reflect partial equivalence.

We report strict accuracy (only score-2 pairs), and lenient accuracy (score-1 and score-2 pairs). Across non-Zh--En language pairs, the LLM-based evaluation yields a mean strict accuracy of 0.685 and a lenient accuracy of 0.923, where each language pair is treated as one observation. The corresponding 95\% confidence intervals are computed using a t-interval over language pairs (strict: [0.645, 0.724]; lenient: [0.907, 0.940]). Details are provided in Appendix~\ref{app:alignment_quality}.

In addition, performance varies substantially across language pairs. High-resource or closely related pairs such as En--De, En--Es, En--Pt, and Pt--Es achieve very high strict accuracy (up to 0.96), with most annotations assigned score~2. In contrast, more distant pairs such as De--Ja, Fr--Ja, and Zh--Es show lower accuracy and a larger proportion of score~1 and score~0 cases, reflecting the greater difficulty of establishing idiomatic equivalence across typologically distant languages.

\subsection{Similarity Characterization}
\label{sec:benchmark_similarity}
Complementing the alignment quality evaluation, we analyze gloss-based similarity scores to understand how embedding-space signals support and characterize the constructed alignments.

\paragraph{Overall distribution.}
Across 18{,}785 aligned pairs, construction-time similarity scores span a wide range (min $=0.368$, max $\approx 1.0$) with moderately high central tendency (mean $=0.670$, median $=0.651$). Near-saturation scores are rare and typically correspond to highly formulaic or nearly identical glosses (see Appendix~\ref{app:qwen-global}).

\paragraph{Threshold sensitivity.}
To assess the effect of encoder calibration on coverage, we sweep nine thresholds $t$ over the shared overlap interval of the two encoders’ score distributions and count pairs with $sim \ge t$ (see Appendix \ref{app:qwen-threshold}). In Figure~\ref{fig:coverage_curves}, both encoders produce monotonic retention curves but diverge substantially at the same absolute threshold, reflecting calibration and scaling differences rather than semantic disagreement. This supports treating similarity as a relative diagnostic signal and motivates distribution-aware filtering.

\begin{figure}[h]
  \centering
\includegraphics[width=0.7\linewidth]{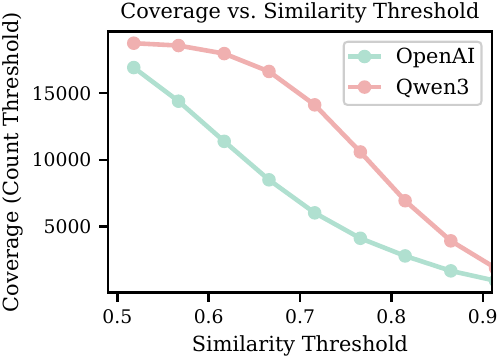} 
\caption{
Retention curves for OpenAI text-embedding-3-large and Qwen3-Embedding-8B as the similarity threshold $t$ varies, illustrating encoder-dependent calibration effects under fixed absolute thresholds.
}
  \label{fig:coverage_curves}
\end{figure}

\paragraph{Embedding consistency and calibration.}
Embedding consistency refers to the agreement in relative similarity structure across different embedding models. When we recompute similarities with independent multilingual encoder, Qwen3-Embedding-8B \cite{qwenembed2025}, we observe a systematic increase in absolute cosine similarity while maintaining strong agreement in relative structure (Pearson $r=0.807$, Spearman $\rho=0.784$; see Appendix~\ref{app:qwen-corr}). This suggests that absolute similarity values are encoder-dependent, while relative similarity structure is largely preserved.

\section{Experiments}
\label{sec:experiments}
We evaluate cross-lingual idiom alignment under two complementary settings.
Task~1 formulates alignment as a controlled multiple-choice problem, enabling fine-grained diagnosis of error patterns through typed distractors.
Task~2 evaluates open-ended target-idiom generation in a large output space and contrasts \textit{No-gloss} and \textit{With-gloss} inputs to assess the effect of an explicit semantic pivot under semantic ambiguity and surface-form mismatch.
These settings support reproducible quantitative comparison and reveal recurring failure modes in cross-lingual idiom alignment.

\begin{table*}[t]
\centering
\small
\begin{tabular}{lcccccccc}
\toprule
\textbf{Model} &
\multicolumn{2}{c}{\textbf{Zh-target}} &
\multicolumn{2}{c}{\textbf{En-target}} &
\multicolumn{2}{c}{\textbf{Other targets}} &
\multicolumn{2}{c}{\textbf{Overall}} \\
\cmidrule(lr){2-3} \cmidrule(lr){4-5} \cmidrule(lr){6-7} \cmidrule(lr){8-9}
& Macro& Micro& Macro& Micro& Macro& Micro& Macro& Micro\\
\midrule
DeepSeek-V3.2 (NT) & 56.84 &  56.96& 57.84 &  56.16& 47.39 &  47.53& 52.70&  53.65\\
DeepSeek-V3.2 (T)    & 69.37 &  69.18& 66.70 &  65.38& 53.92 &  54.09& 61.45&  63.02\\
Gemini-2.5-Pro              & 67.77 &  67.57& 66.53 &  64.95& 62.82 &  62.91& 65.13&  65.17\\
Claude-4.5-Haiku           & 56.80 &  57.30& 55.54 &  53.31& 44.03 &  43.44& 50.50&  51.47\\
Ministral-8B-Instruct       & 28.63 &  28.51& 30.29 &  28.97& 27.78 &  27.76& 28.68&  28.43\\
Qwen3-8B (NT)      & 32.69 &  33.72& 33.95 &  33.62& 24.02 &  23.99& 28.98&  30.56\\
Qwen3-8B (T)         &       41.20&  41.89&       39.57&  39.10&       31.29&  31.39&       36.14&  37.55\\
\bottomrule
\end{tabular}
\caption{Multiple-choice accuracy aggregated by target-language groups. Micro is instance-weighted within each target group (Zh-/En-/Other-target), while Macro averages over directions. All numbers are percentages.}
\label{tab:task1_overall_acc}
\end{table*}

\subsection{Task 1:Multiple-Choice Idiom Equivalence}
\label{sec:task1}
\textbf{Task formulation.}
Given a source-language idiom, the model selects the meaning-equivalent target-language option from a 4-way candidate set. Each instance contains one canonical target idiom from G-IdiomAlign and three typed distractors, enabling error analysis by distractor type in addition to accuracy. To reduce positional bias, we shuffle the option order per instance and store option-type labels (reference, LT, LC, CA; defined below). The model outputs a single choice (A-D) under greedy decoding (temperature $=0$, top-$p=1$). We evaluate 30 direction settings in total, including 16 directions with \emph{high-resource} target languages (Chinese or English; 8 each) and 14 additional directions with other target languages. Here a \emph{direction} is an ordered mapping from a source language to a target language, and we group directions by the target language (Zh-target, En-target, and other targets).

\textbf{Candidate construction.}
The correct (reference) option is the canonical target idiom from G-IdiomAlign. We construct three types of target-language distractors: \textit{Literal Translation Trap} (LT), a word-for-word translation of the source idiom; \textit{Lexical Cue Trap} (LC), a target-language idiom that shares a \emph{partial} lexical cue with the literal translation (typically one salient content word) but conveys an unrelated meaning; and \textit{Contextual Association Trap} (CA), a target-language idiom that is contextually plausible yet semantically opposite. Distractors are generated using Qwen-Max under a unified prompt with explicit type constraints; the details are provided in the Appendix \ref{app:task1_prompt}.

\paragraph{Validity checks for LLM-generated distractors.}
To assess whether LLM-generated distractors introduce superficial shortcuts, we conduct an option-only control and a manual validity check. The results show no significant preference for the gold option over chance and confirm 81.5\% distractor-type validity. Details are provided in Appendix~\ref{app:distractor_validity}.

\subsection{Task 2: Gloss-Contrastive Generation}
\label{sec:task2}

\paragraph{Task formulation.} In the open-ended setting, the model is given a source-language idiom and must generate a meaning-equivalent idiom in the target language. We compare two input conditions: \textit{No-gloss} (source idiom only) and \textit{With-gloss} (source idiom plus its English gloss), where the gloss provides an explicit semantic pivot. We evaluate Task~2 on all 72 directions available in G-IdiomAlign, using greedy decoding (temperature $=0$, top-$p=1$). We require models to output exactly one target-language idiom with no additional explanation, enabling deterministic parsing and automatic scoring (see Appendix~\ref{app:task2_prompts}).

\paragraph{Automatic evaluation.} Because multiple outputs can be valid in open-ended generation, we use an embedding-based semantic matching proxy for coarse-grained scoring. We embed the model output and the canonical target idiom in G-IdiomAlign using Qwen3-Embedding-8B and compute cosine similarity. We report Acc@\(t\), counting a prediction as correct if the similarity exceeds a threshold $t$ in the same embedding space. We choose two operating points, $t=0.70$ and $t=0.80$: $0.70$ is a more permissive threshold, while $0.80$ is a stricter threshold close to the median similarity of canonical aligned pairs in G-IdiomAlign under Qwen3-Embedding-8B (median $\approx 0.78$). This proxy supports aggregate comparison but is not a definitive correctness criterion; for example, it can under-count valid synonymous idioms that diverge from the canonical reference. We therefore treat embedding similarity as a high-confidence semantic indicator rather than a complete estimate of idiom-form correctness, and report a small human evaluation together with auxiliary surface-form metrics (EM/BLEU/ChrF) in Appendix~\ref{app:task2_metric_validity}. 
We interpret Task~2 results primarily as comparative trends.
\subsection{Models}
\label{sec:models}
We evaluate several open-source LLMs and proprietary, including DeepSeek-V3.2 (T/NT) \cite{deepseekai2025deepseekv32pushingfrontieropen}, Gemini-2.5-Pro, Claude-4.5-Haiku, Ministral-8B-Instruct \cite{mistral2024ministral8b}, and Qwen3-8B (T/NT) \cite{yang2025qwen3}. T denotes \emph{Thinking mode} and NT is \emph{No thinking mode}. During dataset preprocessing, we embed glosses with text-embedding-3-large to select high-confidence reference alignments. For Task~1, we generate typed distractors using Qwen-Max \cite{qwenmax2025}. For Task~2 automatic scoring, we use Qwen3-Embedding-8B. We run open-source models on a NVIDIA A40 GPU, while proprietary models are accessed via official APIs.

\begin{table*}[t]
\centering
\small
\begin{tabular}{lccc|ccc|c}
\toprule
& \multicolumn{3}{c}{\textbf{No-gloss}} & \multicolumn{3}{c}{\textbf{With-gloss}} & \\
Model & MeanSim & Acc@0.70 & Acc@0.80 & MeanSim & Acc@0.70 & Acc@0.80 & \(\Delta\)Acc@0.80 \\
\midrule
DeepSeek-V3.2 (NT) & 68.30 & 41.99 & 19.90 & 70.94 & 48.35 & 23.57 & 3.66 \\
DeepSeek-V3.2 (T)  & 72.31 & 49.01 & 24.66 & \textbf{74.24} & 54.90 & \textbf{29.63} & 4.97 \\
Gemini-2.5-Pro     & 68.71 & 44.06 & 19.70 & 72.54 & 51.09 & 27.11 & 7.41 \\
Claude-4.5-Haiku   & 70.90 & 45.47 & 20.47 & 72.94 & 52.16 & 25.66 & 5.19 \\
Ministral-8B-Instruct & 68.30 & 37.97 & 12.96 & 71.69 & 50.00 & 21.22 & 8.28 \\
Qwen3-8B (NT)      & 67.54 & 36.29 & 11.59 & 70.69 & 47.09 & 18.80 & 7.22 \\
Qwen3-8B (T)       & 68.53 & 39.03 & 14.04 & 71.40 & 49.16 & 20.86 & 6.82 \\
\bottomrule
\end{tabular}
\caption{Gloss-Contrastive Generation results. MeanSim ($100\times sim$) and Acc@\(t\) are direction-averaged (macro) over 72 language directions; Acc@\(t\) counts a prediction as correct if cosine \(sim \ge t\). We report \(t=0.70\) and \(t=0.80\). \(\Delta\)Acc@0.80 is the \textit{With-gloss} minus \textit{No-gloss} improvement. Bold indicates the best performance.}
\label{tab:task2_main}
\end{table*}

\begin{figure}[t]
    \centering
    \includegraphics[width=0.95\linewidth]{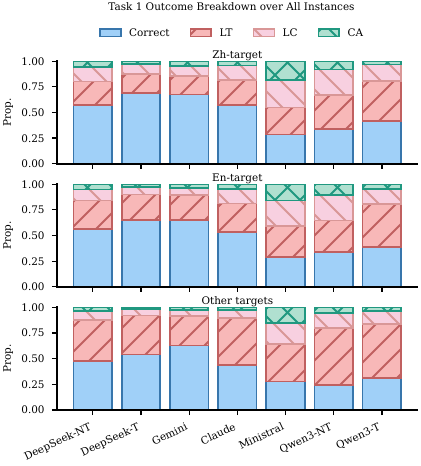}
    \caption{Task~1 outcomes by target-language regime. Stacked bars (normalized within each regime) decompose specific instances into Correct predictions and LT/LC/CA, highlighting differences between high-resource targets (Zh/En) and other target languages.}
    \label{fig:task1_outcome_by_regime}
\end{figure}

\section{Results and Analysis}\label{sec:Result_Analysis}
\subsection{Multiple-Choice Idiom Equivalence}

\subsubsection{Overall Accuracy Across Target Groups}
Table~\ref{tab:task1_overall_acc} reports Task~1 accuracy by target-language group.
Across models, performance is consistently lower on Other targets directions than on Zh-target or En-target, suggesting that idiom alignment is more challenging when the target language is not a high-resource language such as Chinese or English.
This gap is most pronounced for lower-performing models.
Gemini-2.5-Pro achieves the strongest overall performance, particularly on Other targets.
Enabling \emph{thinking mode} yields consistent gains across models, with clear improvements for DeepSeek-V3.2 and Qwen3-8B.

\subsubsection{Outcome Decomposition}
\label{sec:task1_error_types}
Figure~\ref{fig:task1_outcome_by_regime} decomposes Task~1 outcomes by target-language regime into \textit{Correct} predictions and three distractor types (LT/LC/CA); proportions are in Appendix~\ref{app:task1_outcomes}. These patterns are further illustrated with representative examples and error cases in Appendix~\ref{app:task1_casestudy}, which provide concrete instances of LT-, LC-, and CA-driven errors. The Other targets regime has fewer correct predictions, consistent with lower accuracy. 

Across models, \textit{Literal Translation Trap} (LT) dominates errors, especially for lower-resource targets, indicating stronger literal-transfer attraction. 
In contrast, \textit{Zh-target} directions show more \textit{Lexical Cue} (LC) errors, suggesting partial lexical overlap misleads models when Chinese is the target. 
Lower-performing models also exhibit higher \textit{Contextual Association} (CA) rates, particularly in Other targets.

Enabling \emph{thinking mode} improves accuracy for DeepSeek-V3.2 (reducing LT and LC errors) and Qwen3-8B (mainly decreasing LC and CA, with LT still dominant). Overall, literal-translation attraction remains the primary challenge in cross-lingual idiom alignment.

\subsection{Gloss-Contrastive Generation}
\textbf{Overall performance.}
Table~\ref{tab:task2_main} summarizes Task~2 results under the semantic-similarity proxy.
Providing an English gloss (\textit{With-gloss}) improves MeanSim and Acc@\(t\) for every model, with consistent gains at the stricter threshold Acc@0.80, suggesting that glosses help constrain generation toward the intended meaning.
Despite this, Acc@0.80 remains modest even with gloss, highlighting the difficulty of producing meaning-equivalent idioms in an unconstrained output space.

Under \textit{With-gloss}, DeepSeek-V3.2 (T) achieves the strongest overall performance.
For models with both variants, enabling thinking yields consistent improvements, with the advantage more apparent at higher similarity thresholds.
Results at additional thresholds are reported in Appendix~\ref{app:task2_thresholds}. Representative examples and error cases for open-ended generation are provided in Appendix~\ref{app:task2_casestudy}, illustrating both correct outputs and acceptable paraphrases that may be under- or over-estimated by the similarity-based metric.

%\subsection{Attention-based Diagnostics} \label{sec: attention}
%We introduce attention-based diagnostics for Task~2 to characterize how the model allocates attention over \emph{input spans} during decoding and how these allocations relate to output correctness. Throughout, we treat attention strictly as a \textbf{correlational diagnostic signal}. All analyses are conducted on \textbf{Qwen3-8B} with greedy decoding (temperature \(=0\), top-\(p=1\)). Post-softmax self-attention weights are extracted using \texttt{TransformerLens} hooks.\footnote{https://github.com/TransformerLensOrg/TransformerLens} For each input condition, we annotate 200 generations (400 total) and exclude Type~0 invalid outputs (empty, garbled, or not interpretable as meaningful text in the target language), yielding \(n=189\) valid instances in \textit{With-gloss} and \(n=189\) in \textit{No-gloss}. Table~\ref{tab:attn_outcome_counts} reports the outcome breakdown by error type. Two collaborators label outputs independently (Cohen's \(\kappa=0.81\)) and resolve disagreements by discussion. We use the following error taxonomy (Appendix~\ref{app:annotation_scheme}): Type~2 (T2, literal word-by-word translation missing the idiom’s figurative meaning), Type~3 (T3, meaning-correct but non-idiomatic), and Type~4 (T4, meaning-incorrect and not a word-by-word literal translation). 

\subsection{Attention-based Diagnostics}
\label{sec:attention}

We introduce attention-based diagnostics for Task~2 to characterize how the model allocates attention over \emph{input spans} during decoding and how these allocations relate to output quality. Throughout, we treat attention strictly as a \textbf{correlational diagnostic signal}. All analyses are conducted on \textbf{Qwen3-8B} with greedy decoding (temperature \(=0\), top-\(p=1\)). Post-softmax self-attention weights are extracted using \texttt{TransformerLens} hooks\footnote{https://github.com/TransformerLensOrg/TransformerLens}.

For each input condition, we annotate 200 generations (400 total) and exclude Type~0 invalid outputs (empty, garbled, or not interpretable as meaningful target-language text), yielding \(n=189\) valid instances in \textit{With-gloss} and \(n=189\) in \textit{No-gloss}. Table~\ref{tab:attn_outcome_counts} reports the outcome breakdown by error type. Two collaborators independently label the outputs (Cohen's \(\kappa=0.81\)) and resolve disagreements by discussion. We use the following error taxonomy: Type~2 (T2, literal word-by-word translation missing the idiom’s figurative meaning), Type~3 (T3, meaning-correct but non-idiomatic), and Type~4 (T4, meaning-incorrect and not a word-by-word literal translation).

\begin{table}[h]
\centering
\small
\begin{tabular}{lcccc}
\toprule
Condition & Valid & Correct & Wrong & T2 / T3 / T4\\
\midrule
\textit{With-gloss}    & 189 & 123 & 66  & 6 / 30 / 30 \\
\textit{No-gloss} & 189 & 53  & 136 & 60 / 40 / 36 \\
\bottomrule
\end{tabular}
\caption{Outcome composition for the annotated subset (Type~0 excluded). T2/T3/T4 denote the breakdown of error types within Wrong outputs.}
\label{tab:attn_outcome_counts}
\end{table}

%\paragraph{Head/layer-level structural overlap.} To localize cross-condition shifts, we score each attention head by span attention mass averaged over decision positions and select per-example top-\(k\) salient heads (\(k=10\)). We then form condition-level global top-\(m\) head sets by frequency across examples (\(m=50\)) and compute Jaccard overlap between \textit{With-gloss} and \textit{No-gloss}. We additionally derive induced layer sets by projecting the selected heads to their corresponding layers (Apppendix~\ref{app:attn_salience_defs}). As shown in Table~\ref{tab:attn_heads_layers}, head overlap is substantially lower than layer overlap overall (\(J_{\text{heads}}=0.32\) vs.\ \(J_{\text{layers}}=0.90\)). This pattern also holds across subsets, although layer overlap is lower for Correct-only than for Wrong-only outputs. These results suggest that cross-condition differences are expressed more strongly through head-level reconfiguration than broad layer-level shifts: the two conditions largely recruit similar layers, but differ in which heads within those layers are salient. The lower layer overlap for Correct-only outputs further suggests less shared layer-level structure across conditions for correct than for wrong cases.

\paragraph{Head/layer-level structural overlap.}
To assess whether cross-condition differences are driven more by head changes or layer shifts, we compare the cross-condition overlap of the salient heads and salient layers (Table~\ref{tab:attn_heads_layers}; see Appendix~\ref{app:attn_salience_defs} for details).

As shown in Table~\ref{tab:attn_heads_layers}, head overlap is substantially lower than layer overlap overall (\(J_{\text{heads}}=0.32\) vs.\ \(J_{\text{layers}}=0.90\)). This pattern also holds across subsets, although layer overlap is lower for Correct-only than for Wrong-only outputs. These results suggest that cross-condition differences are expressed more strongly through head-level reconfiguration than broad layer-level shifts: the two conditions largely recruit similar layers but differ in which heads within those layers are salient. Lower layer overlap for Correct-only outputs further suggests less shared layer-level structure across conditions for correct than for wrong cases.

\begin{table}[h]
\centering
\small
\begin{tabular}{lcc}
\toprule
Subset & \(J_{\text{heads}}\) & \(J_{\text{layers}}\) \\
\midrule
All         & 0.32& 0.90\\
Correct-only& 0.32& 0.76\\
Wrong-only  & 0.37& 0.90\\
\bottomrule
\end{tabular}
\caption{Cross-condition overlap between \textit{With-gloss} and \textit{No-gloss} for salient head sets and salient layer sets (Jaccard; higher indicates more overlap).}
\label{tab:attn_heads_layers}
\end{table}

\begin{table*}[t]
\centering
\small
\setlength{\tabcolsep}{3pt}
\begin{tabular}{llcccccccc}
\toprule
& & \multicolumn{4}{c}{\textit{With-gloss}} & \multicolumn{4}{c}{\textit{No-gloss}} \\
\cmidrule(lr){3-6}\cmidrule(lr){7-10}
Metric & Analysis 
& Correct & Wrong & $\delta$ & $q$ 
& Correct & Wrong & $\delta$ & $q$ \\
\midrule
IAR & CvsW 
& 0.03 (0.01) & 0.03 (0.01) & -0.10 & 0.250 
& 0.06 (0.02) & 0.06 (0.02) & 0.18 & 0.087 \\
\addlinespace
GAR & CvsW 
& 0.07 (0.04) & 0.06 (0.03) & 0.24 & 0.026 
& -- & -- & -- & -- \\
\addlinespace
DR & CvsW 
& 0.90 (0.04) & 0.91 (0.03) & -0.20 & 0.033 
& 0.94 (0.02) & 0.94 (0.02) & -0.18 & 0.087 \\
\addlinespace
OtherTop1 & CvsW 
& 0.57 (0.03) & 0.58 (0.02) & -0.21 & 0.033 
& 0.58 (0.02) & 0.58 (0.02) & 0.01 & 0.900 \\
\bottomrule
\end{tabular}
\caption{
Token-level attention diagnostics (Correct vs.\ Wrong).
Each metric measures where the model attends during generation:
IAR, GAR, DR, and OtherTop1.
We report median (IQR) for each group, Cliff’s $\delta$, 
and $q$-values from two-sided Mann-Whitney U tests with BH-FDR correction within each condition.
}
\label{tab:attn_cvsw}
\end{table*}

\begin{table*}[t]
\centering
\small
\setlength{\tabcolsep}{3pt}
\begin{tabular}{llcccccccc}
\toprule
& & \multicolumn{4}{c}{\textit{With-gloss}} & \multicolumn{4}{c}{\textit{No-gloss}} \\
\cmidrule(lr){3-6}\cmidrule(lr){7-10}
Metric & Analysis 
& Type 2 & Type 3 & Type 4 & $q$ 
& Type 2 & Type 3 & Type 4 & $q$ \\
\midrule
IAR & WrongType 
& 0.03 (0.03) & 0.03 (0.02) & 0.03 (0.01) & 0.878 
& 0.05 (0.03) & 0.06 (0.02) & 0.05 (0.02) & $<0.001$ \\
\addlinespace
GAR & WrongType 
& 0.05 (0.01) & 0.07 (0.03) & 0.06 (0.04) & 0.269 
& -- & -- & -- & -- \\
\addlinespace
DR & WrongType 
& 0.91 (0.03) & 0.90 (0.03) & 0.91 (0.03) & 0.269 
& 0.95 (0.03) & 0.94 (0.02) & 0.95 (0.02) & $<0.001$ \\
\addlinespace
OtherTop1 & WrongType 
& 0.58 (0.02) & 0.58 (0.02) & 0.58 (0.02) & 0.269 
& 0.58 (0.02) & 0.57 (0.03) & 0.58 (0.02) & $<0.001$ \\
\bottomrule
\end{tabular}
\caption{
Token-level attention diagnostics across error types (Type 2/3/4).
We report median (IQR) per error type and $q$-values from Kruskal-Wallis tests 
with BH-FDR correction within each condition.
}
\label{tab:attn_wrongtype}
\end{table*}

\paragraph{Token-level diagnostics.}
Let \(\bar{A}(k)\) denote the aggregated post-softmax attention mass assigned to input key
token \(k\), computed by averaging attention over layers and heads and over
template-defined generation positions corresponding to the content-bearing segment.
Here \(k\) ranges over key-token positions in the full prompt sequence.
We interpret \(\bar{A}(k)\) as the model’s average attention mass assigned to token \(k\) during the generation of the content-bearing output (see Appendix~\ref{app:attn_token_defs}). 

Using tokenizer offset mapping, we map the idiom span and (when available) the gloss span
to token-index sets \(K_{\text{idiom}}\) and \(K_{\text{gloss}}\) in the full prompt sequence.
We then define:

\[
\mathrm{IAR}=\sum_{k\in K_{\text{idiom}}}\bar{A}(k)
\]
\(\mathrm{IAR}\) (Idiom Attention Ratio) is the total attention mass on the idiom span; larger
values indicate stronger concentration on idiom tokens.

\[
\mathrm{GAR}=\sum_{k\in K_{\text{gloss}}}\bar{A}(k)\ \ (\textit{With-gloss})
\]
\(\mathrm{GAR}\) (Gloss Attention Ratio) is the total attention mass on the gloss span
(defined only under \textit{With-gloss}); larger values indicate stronger anchoring to the
provided gloss.

\[
\mathrm{DR}=1-\mathrm{IAR}-\mathrm{GAR}
\]
\(\mathrm{DR}\) (Diffuse Ratio) captures the residual attention mass outside the
tracked spans; larger values indicate greater allocation to other context tokens.
Under \textit{No-gloss}, \(\mathrm{GAR}\) is defined as $0$, so \(\mathrm{DR}=1-\mathrm{IAR}\).

\[
\mathrm{OtherTop1}
=
\max_{k \in K_{\mathrm{other}}}
\frac{\bar{A}(k)}{\sum_{j \in K_{\mathrm{other}}}\bar{A}(j)}
\]

\(\mathrm{OtherTop1}\) is the maximum share among off-span tokens after renormalizing within the off-span set; higher values indicate a stronger off-span peak.

For Correct-vs.-Wrong comparisons, we use two-sided Mann-Whitney U tests,
a non-parametric two-group comparison, and report Cliff's \(\delta\), whose sign indicates
the direction of the difference and whose magnitude reflects its strength.
For comparisons across Types~2/3/4 within wrong outputs, we use Kruskal-Wallis tests,
which assess whether the error types differ overall without assuming normality.
Within each condition, \(p\)-values are adjusted across metric-wise tests using BH-FDR. Table~\ref{tab:attn_cvsw} and \ref{tab:attn_wrongtype} summarize the result of token-level diagnostics.

\textbf{\emph{With-gloss}} (Type~0 excluded; \(n=189\)).
Correct outputs exhibit stronger gloss anchoring and reduced off-span allocation:
GAR increases, while both DR and OtherTop1 decrease (all \(q<0.05\)).
By contrast, idiom-span mass does not distinguish Correct from Wrong (IAR; \(q=0.250\)).
Within wrong outputs, cross-type differences are not robust after correction.

\textbf{\emph{No-gloss}} (Type~0 excluded; \(n=189\); GAR not applicable).
The Correct-vs.-Wrong contrast is weaker and does not survive correction for IAR or DR,
and OtherTop1 shows no meaningful difference.
Nevertheless, WrongType comparisons are strongly structured by error type across the
available diagnostics (IAR/DR/OtherTop1; all \(q<0.001\)),
suggesting more heterogeneous failure modes in the absence of an explicit gloss anchor.

Overall, glosses consistently improve open-ended idiom generation, and attention diagnostics
suggest that correctness under \textit{With-gloss} aligns with stronger gloss anchoring,
whereas \textit{No-gloss} errors exhibit more heterogeneous attention patterns across error types.

\section{Conclusion}
We present \textbf{G-IdiomAlign}, a gloss-pivoted benchmark supporting {Multiple-Choice Idiom Equivalence} and {Gloss-Contrastive Generation} to diagnose literal-translation biases across LLMs.
% We presented \textbf{G-IdiomAlign}, a gloss-pivoted benchmark designed to diagnose cross-lingual idiom alignment under controlled yet challenging evaluation settings. 
% By combining a structured alignment task with an open-ended generation task, the benchmark exposes systematic failure patterns that are difficult to detect with unconstrained or purely accuracy-driven evaluations.
Our results, spanning diverse proprietary and open-source models, highlight literal-translation attraction as a persistent obstacle in cross-lingual idiom alignment.
% While providing explicit glosses consistently improves open-ended generation under an embedding-based semantic proxy, performance remains far from saturated. This underscores the difficulty of producing meaning-equivalent idioms, motivating further developments in robust idiom translation.
Attention-based diagnostics further suggest that successful \textit{With-gloss} generations are associated with stronger anchoring to gloss information.
% with cross-condition differences manifesting prominently at the level of attention heads.
% While providing an explicit gloss consistently improves open-ended generation under an embedding-based semantic proxy, overall performance remains far from saturated, underscoring the difficulty of producing meaning-equivalent idioms in an unconstrained output space, motivating further developments in robust idiom translation.
While providing explicit glosses consistently improves open-ended generation under an embedding-based semantic proxy, performance remains far from saturated. This motivates further developments in robust idiom translation.

\section*{Limitations}
G-IdiomAlign is designed as a precision-first benchmark for diagnosing cross-lingual idiom alignment rather than exhaustive idiomatic equivalence. This improves interpretability and reproducibility, but limits coverage and external validity.

\textbf{English-pivot bias.}
We use English Wiktionary glosses as a single semantic pivot across nine languages. This improves consistency, but may introduce English-centric bias because glosses can compress pragmatic or culture-specific meaning and vary in style and granularity across editions.

\textbf{Trade-offs in reference alignments.}
To ensure unambiguous supervision, we apply single-sense filtering and exclude polysemous idioms. This improves interpretability but removes sense selection and reduces coverage. We further impose a one-to-one constraint via MNN, which favors high-precision pairs but under-represents many-to-many relations such as synonym clusters. 

\textbf{Dependence on embeddings, proxies, and tools.}
The pipeline relies on embedding-based retrieval and filtering, as well as an LLM for distractor generation. As a result, benchmark construction is sensitive to embedding calibration, dataset size, and gloss noise, and may inherit model-specific biases. Task~2 further uses fixed-threshold embedding similarity as a scalable proxy, which may miss valid non-canonical generations and may not be fully comparable across languages, such dependence remains a limitation.

\textbf{Limited generality.}
Our attention analyses are correlational rather than causal, and are based on a single model with a modest annotated sample under greedy decoding. The observed patterns may therefore not generalize across models, decoding settings, or language directions.

\section*{Ethical Considerations}
This work involves several value-sensitive design choices. First, we use English glosses as a shared semantic pivot to enable controlled cross-lingual idiom alignment. While this results in high-confidence alignment, it may introduce English-centric bias and compress culture-specific pragmatic or stylistic distinctions encoded in non-English idioms. We treat this as a deliberate trade-off for diagnostic clarity, rather than as a claim of cultural neutrality.

Second, idioms are culturally grounded expressions, and operationalizing idiomatic equivalence through glosses and embedding-based similarity necessarily abstracts away contextual and sociocultural nuance. Our benchmark is therefore intended to support analysis of model behavior under controlled conditions, not to define authoritative judgments of idiomatic correctness across cultures.

Finally, our evaluation metrics, especially the embedding-based proxy in Gloss-Contrastive Generation, are designed for consistent comparison rather than deployment. We caution against using benchmark scores as standalone indicators of translation quality or fairness in real-world applications.


\section*{Acknowledgments}
This work was supported in part by the Science and Technology Development Fund of Macau SAR (Grant Nos. FDCT/0007/2024/AKP, EF2024-00185-FST), the UM and UMDF (Grant Nos. MYRG-GRG2024-00165-FST-UMDF, MYRG-GRG2025-00236-FST), the Tencent AI Lab Rhino-Bird Research Program (Grant No. EF2023-00151-FST), the Stanley Ho Medical Development Foundation (Grant No. SHMDF-AI/2026/001), and the National Natural Science Foundation of China (Grant No. 62266013).

\bibliography{custom}

@inproceedings{haagsma-etal-2020-magpie,
    title = "{MAGPIE}: A Large Corpus of Potentially Idiomatic Expressions",
    author = "Haagsma, Hessel  and
      Bos, Johan  and
      Nissim, Malvina",
    editor = "Calzolari, Nicoletta  and
      B{\'e}chet, Fr{\'e}d{\'e}ric  and
      Blache, Philippe  and
      Choukri, Khalid  and
      Cieri, Christopher  and
      Declerck, Thierry  and
      Goggi, Sara  and
      Isahara, Hitoshi  and
      Maegaard, Bente  and
      Mariani, Joseph  and
      Mazo, H{\'e}l{\`e}ne  and
      Moreno, Asuncion  and
      Odijk, Jan  and
      Piperidis, Stelios",
    booktitle = "Proceedings of the Twelfth Language Resources and Evaluation Conference",
    month = may,
    year = "2020",
    address = "Marseille, France",
    publisher = "European Language Resources Association",
    url = "https://aclanthology.org/2020.lrec-1.35/",
    pages = "279--287",
    language = "eng",
    ISBN = "979-10-95546-34-4"
}

@inproceedings{mi-etal-2025-rolling,
    title = "Rolling the {DICE} on Idiomaticity: How {LLM}s Fail to Grasp Context",
    author = "Mi, Maggie  and
      Villavicencio, Aline  and
      Moosavi, Nafise Sadat",
    editor = "Che, Wanxiang  and
      Nabende, Joyce  and
      Shutova, Ekaterina  and
      Pilehvar, Mohammad Taher",
    booktitle = "Proceedings of the 63rd Annual Meeting of the Association for Computational Linguistics (Volume 1: Long Papers)",
    month = jul,
    year = "2025",
    address = "Vienna, Austria",
    publisher = "Association for Computational Linguistics",
    url = "https://aclanthology.org/2025.acl-long.362/",
    doi = "10.18653/v1/2025.acl-long.362",
    pages = "7314--7332",
    ISBN = "979-8-89176-251-0"
}

@inproceedings{sentsova-etal-2025-multicopie,
    title = "{M}ulti{C}o{PIE}: A Multilingual Corpus of Potentially Idiomatic Expressions for Cross-lingual {PIE} Disambiguation",
    author = "Sentsova, Uliana  and
      Ciminari, Debora  and
      Genabith, Josef Van  and
      Espa{\~n}a-Bonet, Cristina",
    editor = {Ojha, Atul Kr.  and
      Giouli, Voula  and
      Mititelu, Verginica Barbu  and
      Constant, Mathieu  and
      Korvel, Gra{\v{z}}ina  and
      Do{\u{g}}ru{\"o}z, A. Seza  and
      Rademaker, Alexandre},
    booktitle = "Proceedings of the 21st Workshop on Multiword Expressions (MWE 2025)",
    month = may,
    year = "2025",
    address = "Albuquerque, New Mexico, U.S.A.",
    publisher = "Association for Computational Linguistics",
    url = "https://aclanthology.org/2025.mwe-1.8/",
    doi = "10.18653/v1/2025.mwe-1.8",
    pages = "67--81",
    ISBN = "979-8-89176-243-5"
}

@misc{saxena2020epie,
      title={{EPIE Dataset: A Corpus For Possible Idiomatic Expressions}}, 
      author={Prateek Saxena and Soma Paul},
      year={2020},
      eprint={2006.09479},
      archivePrefix={arXiv},
      primaryClass={cs.CL}
}

@inproceedings{de-luca-fornaciari-etal-2024-hard,
    title = "{A Hard Nut to Crack: Idiom Detection with Conversational Large Language Models}",
    author = "De Luca Fornaciari, Francesca  and
      Altuna, Bego{\~n}a  and
      Gonzalez-Dios, Itziar  and
      Melero, Maite",
    editor = "Ghosh, Debanjan  and
      Muresan, Smaranda  and
      Feldman, Anna  and
      Chakrabarty, Tuhin  and
      Liu, Emmy",
    booktitle = "Proceedings of the 4th Workshop on Figurative Language Processing (FigLang 2024)",
    month = jun,
    year = "2024",
    address = "Mexico City, Mexico (Hybrid)",
    publisher = "Association for Computational Linguistics",
    url = "https://aclanthology.org/2024.figlang-1.5/",
    doi = "10.18653/v1/2024.figlang-1.5",
    pages = "35--44"
}

@inproceedings{tedeschi-etal-2022-id10m,
    title = "{ID}10{M}: Idiom Identification in 10 Languages",
    author = "Tedeschi, Simone  and
      Martelli, Federico  and
      Navigli, Roberto",
    editor = "Carpuat, Marine  and
      de Marneffe, Marie-Catherine  and
      Meza Ruiz, Ivan Vladimir",
    booktitle = "Findings of the Association for Computational Linguistics: NAACL 2022",
    month = jul,
    year = "2022",
    address = "Seattle, United States",
    publisher = "Association for Computational Linguistics",
    url = "https://aclanthology.org/2022.findings-naacl.208/",
    doi = "10.18653/v1/2022.findings-naacl.208",
    pages = "2715--2726"
}

@inproceedings{moussallem-etal-2018-lidioms,
    title = "{LI}dioms: A Multilingual Linked Idioms Data Set",
    author = "Moussallem, Diego  and
      Sherif, Mohamed Ahmed  and
      Esteves, Diego  and
      Zampieri, Marcos  and
      Ngonga Ngomo, Axel-Cyrille",
    editor = "Calzolari, Nicoletta  and
      Choukri, Khalid  and
      Cieri, Christopher  and
      Declerck, Thierry  and
      Goggi, Sara  and
      Hasida, Koiti  and
      Isahara, Hitoshi  and
      Maegaard, Bente  and
      Mariani, Joseph  and
      Mazo, H{\'e}l{\`e}ne  and
      Moreno, Asuncion  and
      Odijk, Jan  and
      Piperidis, Stelios  and
      Tokunaga, Takenobu",
    booktitle = "Proceedings of the Eleventh International Conference on Language Resources and Evaluation ({LREC} 2018)",
    month = may,
    year = "2018",
    address = "Miyazaki, Japan",
    publisher = "European Language Resources Association (ELRA)",
    url = "https://aclanthology.org/L18-1392/"
}

@inproceedings{fu-etal-2025-chengyu,
    title = "{CHENGYU}-{BENCH}: Benchmarking Large Language Models for {C}hinese Idiom Understanding and Use",
    author = "Fu, Yicheng  and
      Huang, Zhemin  and
      Yang, Liuxin  and
      Lu, Yumeng  and
      Dai, Zhongdongming",
    editor = "Christodoulopoulos, Christos  and
      Chakraborty, Tanmoy  and
      Rose, Carolyn  and
      Peng, Violet",
    booktitle = "Proceedings of the 2025 Conference on Empirical Methods in Natural Language Processing",
    month = nov,
    year = "2025",
    address = "Suzhou, China",
    publisher = "Association for Computational Linguistics",
    url = "https://aclanthology.org/2025.emnlp-main.119/",
    doi = "10.18653/v1/2025.emnlp-main.119",
    pages = "2355--2366",
    ISBN = "979-8-89176-332-6"
}

@inproceedings{donthi-etal-2025-improving,
    title = "Improving {LLM} Abilities in Idiomatic Translation",
    author = "Donthi, Sundesh  and
      Spencer, Maximilian  and
      Patel, Om B.  and
      Doh, Joon Young  and
      Rodan, Eid  and
      Zhu, Kevin  and
      O{'}Brien, Sean",
    editor = "Hettiarachchi, Hansi  and
      Ranasinghe, Tharindu  and
      Rayson, Paul  and
      Mitkov, Ruslan  and
      Gaber, Mohamed  and
      Premasiri, Damith  and
      Tan, Fiona Anting  and
      Uyangodage, Lasitha",
    booktitle = "Proceedings of the First Workshop on Language Models for Low-Resource Languages",
    month = jan,
    year = "2025",
    address = "Abu Dhabi, United Arab Emirates",
    publisher = "Association for Computational Linguistics",
    url = "https://aclanthology.org/2025.loreslm-1.13/",
    pages = "175--181"
}

@article{Yang2025EvaluatingLO,
  title={Evaluating {LLMs} on Chinese Idiom Translation},
  author={Cai Yang and Yao Dou and David Heineman and Xiaofeng Wu and Wei Xu},
  journal={ArXiv},
  year={2025},
  volume={abs/2508.10421},
  url={https://api.semanticscholar.org/CorpusID:280649964}
}

@inproceedings{qian-2024-automating,
    title = "Automating Idiom Translation with Cross-Lingual Natural Language Generation Grounded In Semantic Analyses Using Large Language Models",
    author = "Qian, Ming",
    editor = "Martindale, Marianna  and
      Campbell, Janice  and
      Savenkov, Konstantin  and
      Goel, Shivali",
    booktitle = "Proceedings of the 16th Conference of the Association for Machine Translation in the Americas (Volume 2: Presentations)",
    month = sep,
    year = "2024",
    address = "Chicago, USA",
    publisher = "Association for Machine Translation in the Americas",
    url = "https://aclanthology.org/2024.amta-presentations.7/",
    pages = "95--115"
}

@inproceedings{liu-etal-2023-crossing,
    title = "Crossing the Threshold: Idiomatic Machine Translation through Retrieval Augmentation and Loss Weighting",
    author = "Liu, Emmy  and
      Chaudhary, Aditi  and
      Neubig, Graham",
    editor = "Bouamor, Houda  and
      Pino, Juan  and
      Bali, Kalika",
    booktitle = "Proceedings of the 2023 Conference on Empirical Methods in Natural Language Processing",
    month = dec,
    year = "2023",
    address = "Singapore",
    publisher = "Association for Computational Linguistics",
    url = "https://aclanthology.org/2023.emnlp-main.933/",
    doi = "10.18653/v1/2023.emnlp-main.933",
    pages = "15095--15111"
}

@article{Li_Chen_Yuan_Wu_Yang_Tao_Xiao_2024, title={Translate Meanings, Not Just Words: {IdiomKB’s} Role in Optimizing Idiomatic Translation with Language Models}, volume={38}, url={https://ojs.aaai.org/index.php/AAAI/article/view/29817}, DOI={10.1609/aaai.v38i17.29817}, abstractNote={To translate well, machine translation (MT) systems and general-purposed language models (LMs) need a deep understanding of both source and target languages and cultures. Therefore, idioms, with their non-compositional nature, pose particular challenges for Transformer-based systems, as literal translations often miss the intended meaning. Traditional methods, which replace idioms using existing knowledge bases (KBs), often lack scale and context-awareness. Addressing these challenges, our approach prioritizes context-awareness and scalability, allowing for offline storage of idioms in a manageable KB size. This ensures efficient serving with smaller models and provides a more comprehensive understanding of idiomatic expressions. We introduce a multilingual idiom KB (IdiomKB) developed using large LMs to address this. This KB facilitates better translation by smaller models, such as BLOOMZ (7.1B), Alpaca (7B), and InstructGPT (6.7B), by retrieving idioms’ figurative meanings. We present a novel, GPT-4-powered metric for human-aligned evaluation, demonstrating that IdiomKB considerably boosts model performance. Human evaluations further validate our KB’s quality.}, number={17}, journal={Proceedings of the AAAI Conference on Artificial Intelligence}, author={Li, Shuang and Chen, Jiangjie and Yuan, Siyu and Wu, Xinyi and Yang, Hao and Tao, Shimin and Xiao, Yanghua}, year={2024}, month={Mar.}, pages={18554-18563} }

@inproceedings{khoshtab-etal-2025-comparative,
    title = "Comparative Study of Multilingual Idioms and Similes in Large Language Models",
    author = "Khoshtab, Paria  and
      Namazifard, Danial  and
      Masoudi, Mostafa  and
      Akhgary, Ali  and
      Mahdizadeh Sani, Samin  and
      Yaghoobzadeh, Yadollah",
    editor = "Rambow, Owen  and
      Wanner, Leo  and
      Apidianaki, Marianna  and
      Al-Khalifa, Hend  and
      Eugenio, Barbara Di  and
      Schockaert, Steven",
    booktitle = "Proceedings of the 31st International Conference on Computational Linguistics",
    month = jan,
    year = "2025",
    address = "Abu Dhabi, UAE",
    publisher = "Association for Computational Linguistics",
    url = "https://aclanthology.org/2025.coling-main.580/",
    pages = "8680--8698"
}

@inproceedings{li-etal-2023-bilingual,
    title = "On Bilingual Lexicon Induction with Large Language Models",
    author = "Li, Yaoyiran  and
      Korhonen, Anna  and
      Vuli{\'c}, Ivan",
    editor = "Bouamor, Houda  and
      Pino, Juan  and
      Bali, Kalika",
    booktitle = "Proceedings of the 2023 Conference on Empirical Methods in Natural Language Processing",
    month = dec,
    year = "2023",
    address = "Singapore",
    publisher = "Association for Computational Linguistics",
    url = "https://aclanthology.org/2023.emnlp-main.595/",
    doi = "10.18653/v1/2023.emnlp-main.595",
    pages = "9577--9599"
}

@article{Ding_Cao_Zhao_2024, title={Enhancing Bilingual Lexicon Induction via Bi-directional Translation Pair Retrieving}, volume={38}, url={https://ojs.aaai.org/index.php/AAAI/article/view/29744}, DOI={10.1609/aaai.v38i16.29744}, abstractNote={Most Bilingual Lexicon Induction (BLI) methods retrieve word translation pairs by finding the closest target word for a given source word based on cross-lingual word embeddings (WEs). However, we find that solely retrieving translation from the source-to-target perspective leads to some false positive translation pairs, which significantly harm the precision of BLI. To address this problem, we propose a novel and effective method to improve translation pair retrieval in cross-lingual WEs. Specifically, we consider both source-side and target-side perspectives throughout the retrieval process to alleviate false positive word pairings that emanate from a single perspective. On a benchmark dataset of BLI, our proposed method achieves competitive performance compared to existing state-of-the-art (SOTA) methods. It demonstrates effectiveness and robustness across six experimental languages, including similar language pairs and distant language pairs, under both supervised and unsupervised settings.}, number={16}, journal={Proceedings of the AAAI Conference on Artificial Intelligence}, author={Ding, Qiuyu and Cao, Hailong and Zhao, Tiejun}, year={2024}, month={Mar.}, pages={17898-17906} }

@article{he-etal-2025-investigating,
    title = "Investigating Idiomaticity in Word Representations",
    author = "He, Wei  and
      Vieira, Tiago Kramer  and
      Garcia, Marcos  and
      Scarton, Carolina  and
      Idiart, Marco  and
      Villavicencio, Aline",
    journal = "Computational Linguistics",
    volume = "51",
    month = jun,
    year = "2025",
    address = "Cambridge, MA",
    publisher = "MIT Press",
    url = "https://aclanthology.org/2025.cl-2.4/",
    doi = "10.1162/coli_a_00546",
    pages = "505--555"
}

@inproceedings{fakharian-cook-2021-contextualized,
    title = "Contextualized Embeddings Encode Monolingual and Cross-lingual Knowledge of Idiomaticity",
    author = "Fakharian, Samin  and
      Cook, Paul",
    editor = "Cook, Paul  and
      Mitrovi{\'c}, Jelena  and
      Escart{\'i}n, Carla Parra  and
      Vaidya, Ashwini  and
      Osenova, Petya  and
      Taslimipoor, Shiva  and
      Ramisch, Carlos",
    booktitle = "Proceedings of the 17th Workshop on Multiword Expressions (MWE 2021)",
    month = aug,
    year = "2021",
    address = "Online",
    publisher = "Association for Computational Linguistics",
    url = "https://aclanthology.org/2021.mwe-1.4/",
    doi = "10.18653/v1/2021.mwe-1.4",
    pages = "23--32"
}

@inproceedings{zhou-etal-2023-clcl,
    title = "{CLCL}: Non-compositional Expression Detection with Contrastive Learning and Curriculum Learning",
    author = "Zhou, Jianing  and
      Zeng, Ziheng  and
      Bhat, Suma",
    editor = "Rogers, Anna  and
      Boyd-Graber, Jordan  and
      Okazaki, Naoaki",
    booktitle = "Proceedings of the 61st Annual Meeting of the Association for Computational Linguistics (Volume 1: Long Papers)",
    month = jul,
    year = "2023",
    address = "Toronto, Canada",
    publisher = "Association for Computational Linguistics",
    url = "https://aclanthology.org/2023.acl-long.43/",
    doi = "10.18653/v1/2023.acl-long.43",
    pages = "730--743"
}

@inproceedings{zhou-etal-2021-pie,
    title = "{PIE}: A Parallel Idiomatic Expression Corpus for Idiomatic Sentence Generation and Paraphrasing",
    author = "Zhou, Jianing  and
      Gong, Hongyu  and
      Bhat, Suma",
    editor = "Cook, Paul  and
      Mitrovi{\'c}, Jelena  and
      Escart{\'i}n, Carla Parra  and
      Vaidya, Ashwini  and
      Osenova, Petya  and
      Taslimipoor, Shiva  and
      Ramisch, Carlos",
    booktitle = "Proceedings of the 17th Workshop on Multiword Expressions (MWE 2021)",
    month = aug,
    year = "2021",
    address = "Online",
    publisher = "Association for Computational Linguistics",
    url = "https://aclanthology.org/2021.mwe-1.5/",
    doi = "10.18653/v1/2021.mwe-1.5",
    pages = "33--48"
}

@misc{deepseekai2025deepseekv32pushingfrontieropen,
      title={{DeepSeek-V3.2: Pushing the Frontier of Open Large Language Models}}, 
      author={DeepSeek-AI and Aixin Liu and Aoxue Mei and Bangcai Lin and Bing Xue and Bingxuan Wang and Bingzheng Xu and Bochao Wu and Bowei Zhang and Chaofan Lin and Chen Dong and Chengda Lu and Chenggang Zhao and Chengqi Deng and Chenhao Xu and Chong Ruan and Damai Dai and Daya Guo and Dejian Yang and Deli Chen and Erhang Li and Fangqi Zhou and Fangyun Lin and Fucong Dai and Guangbo Hao and Guanting Chen and Guowei Li and H. Zhang and Hanwei Xu and Hao Li and Haofen Liang and Haoran Wei and Haowei Zhang and Haowen Luo and Haozhe Ji and Honghui Ding and Hongxuan Tang and Huanqi Cao and Huazuo Gao and Hui Qu and Hui Zeng and Jialiang Huang and Jiashi Li and Jiaxin Xu and Jiewen Hu and Jingchang Chen and Jingting Xiang and Jingyang Yuan and Jingyuan Cheng and Jinhua Zhu and Jun Ran and Junguang Jiang and Junjie Qiu and Junlong Li and Junxiao Song and Kai Dong and Kaige Gao and Kang Guan and Kexin Huang and Kexing Zhou and Kezhao Huang and Kuai Yu and Lean Wang and Lecong Zhang and Lei Wang and Liang Zhao and Liangsheng Yin and Lihua Guo and Lingxiao Luo and Linwang Ma and Litong Wang and Liyue Zhang and M. S. Di and M. Y Xu and Mingchuan Zhang and Minghua Zhang and Minghui Tang and Mingxu Zhou and Panpan Huang and Peixin Cong and Peiyi Wang and Qiancheng Wang and Qihao Zhu and Qingyang Li and Qinyu Chen and Qiushi Du and Ruiling Xu and Ruiqi Ge and Ruisong Zhang and Ruizhe Pan and Runji Wang and Runqiu Yin and Runxin Xu and Ruomeng Shen and Ruoyu Zhang and S. H. Liu and Shanghao Lu and Shangyan Zhou and Shanhuang Chen and Shaofei Cai and Shaoyuan Chen and Shengding Hu and Shengyu Liu and Shiqiang Hu and Shirong Ma and Shiyu Wang and Shuiping Yu and Shunfeng Zhou and Shuting Pan and Songyang Zhou and Tao Ni and Tao Yun and Tian Pei and Tian Ye and Tianyuan Yue and Wangding Zeng and Wen Liu and Wenfeng Liang and Wenjie Pang and Wenjing Luo and Wenjun Gao and Wentao Zhang and Xi Gao and Xiangwen Wang and Xiao Bi and Xiaodong Liu and Xiaohan Wang and Xiaokang Chen and Xiaokang Zhang and Xiaotao Nie and Xin Cheng and Xin Liu and Xin Xie and Xingchao Liu and Xingkai Yu and Xingyou Li and Xinyu Yang and Xinyuan Li and Xu Chen and Xuecheng Su and Xuehai Pan and Xuheng Lin and Xuwei Fu and Y. Q. Wang and Yang Zhang and Yanhong Xu and Yanru Ma and Yao Li and Yao Li and Yao Zhao and Yaofeng Sun and Yaohui Wang and Yi Qian and Yi Yu and Yichao Zhang and Yifan Ding and Yifan Shi and Yiliang Xiong and Ying He and Ying Zhou and Yinmin Zhong and Yishi Piao and Yisong Wang and Yixiao Chen and Yixuan Tan and Yixuan Wei and Yiyang Ma and Yiyuan Liu and Yonglun Yang and Yongqiang Guo and Yongtong Wu and Yu Wu and Yuan Cheng and Yuan Ou and Yuanfan Xu and Yuduan Wang and Yue Gong and Yuhan Wu and Yuheng Zou and Yukun Li and Yunfan Xiong and Yuxiang Luo and Yuxiang You and Yuxuan Liu and Yuyang Zhou and Z. F. Wu and Z. Z. Ren and Zehua Zhao and Zehui Ren and Zhangli Sha and Zhe Fu and Zhean Xu and Zhenda Xie and Zhengyan Zhang and Zhewen Hao and Zhibin Gou and Zhicheng Ma and Zhigang Yan and Zhihong Shao and Zhixian Huang and Zhiyu Wu and Zhuoshu Li and Zhuping Zhang and Zian Xu and Zihao Wang and Zihui Gu and Zijia Zhu and Zilin Li and Zipeng Zhang and Ziwei Xie and Ziyi Gao and Zizheng Pan and Zongqing Yao and Bei Feng and Hui Li and J. L. Cai and Jiaqi Ni and Lei Xu and Meng Li and Ning Tian and R. J. Chen and R. L. Jin and S. S. Li and Shuang Zhou and Tianyu Sun and X. Q. Li and Xiangyue Jin and Xiaojin Shen and Xiaosha Chen and Xinnan Song and Xinyi Zhou and Y. X. Zhu and Yanping Huang and Yaohui Li and Yi Zheng and Yuchen Zhu and Yunxian Ma and Zhen Huang and Zhipeng Xu and Zhongyu Zhang and Dongjie Ji and Jian Liang and Jianzhong Guo and Jin Chen and Leyi Xia and Miaojun Wang and Mingming Li and Peng Zhang and Ruyi Chen and Shangmian Sun and Shaoqing Wu and Shengfeng Ye and T. Wang and W. L. Xiao and Wei An and Xianzu Wang and Xiaowen Sun and Xiaoxiang Wang and Ying Tang and Yukun Zha and Zekai Zhang and Zhe Ju and Zhen Zhang and Zihua Qu},
      year={2025},
      eprint={2512.02556},
      archivePrefix={arXiv},
      primaryClass={cs.CL},
      url={https://arxiv.org/abs/2512.02556}, 
}

@article{geminiteam2025gemini25,
  title   = {Gemini 2.5: Pushing the Frontier with Advanced Reasoning, Multimodality, Long Context, and Next Generation Agentic Capabilities},
  author  = {Gemini Team, Google},
  year    = {2025},
  journal = {arXiv preprint arXiv:2507.06261},
  url     = {https://arxiv.org/abs/2507.06261}
}

@misc{anthropic2025claudehaiku45,
  title        = {{Claude Haiku 4.5}},
  author       = {Anthropic},
  year         = {2025},
  url = {https://www.anthropic.com/claude/haiku},
}

@misc{mistral2024ministral8b,
  title={Un Ministral, des Ministraux},
  author={{Mistral AI team}},
  year={2024},
  url={https://mistral.ai/news/ministraux/}
}

@misc{yang2025qwen3,
  title        = {Qwen3 Technical Report},
  author       = {Yang, An and Li, Anfeng and Yang, Baosong and Zhang, Beichen and Hui, Binyuan and Zheng, Bo and Yu, Bowen and Gao, Chang and Huang, Chengen and Lv, Chenxu and Zheng, Chujie and Liu, Dayiheng and Zhou, Fan and Huang, Fei and Hu, Feng and Ge, Hao and Wei, Haoran and Lin, Huan and Tang, Jialong and Yang, Jian and Tu, Jianhong and Zhang, Jianwei and Yang, Jianxin and Yang, Jiaxi and Zhou, Jing and Zhou, Jingren and Lin, Junyang and Dang, Kai and Bao, Keqin and Yang, Kexin and Yu, Le and Deng, Lianghao and Li, Mei and Xue, Mingfeng and Li, Mingze and Zhang, Pei and Wang, Peng and Zhu, Qin and Men, Rui and Gao, Ruize and Liu, Shixuan and Luo, Shuang and Li, Tianhao and Tang, Tianyi and Yin, Wenbiao and Ren, Xingzhang and Wang, Xinyu and Zhang, Xinyu and Ren, Xuancheng and Fan, Yang and Su, Yang and Zhang, Yichang and Zhang, Yinger and Wan, Yu and Liu, Yuqiong and Wang, Zekun and Cui, Zeyu and Zhang, Zhenru and Zhou, Zhipeng and Qiu, Zihan},
  year         = {2025},
  eprint       = {2505.09388},
  archivePrefix= {arXiv},
  primaryClass = {cs.CL},
  url          = {https://arxiv.org/abs/2505.09388}
}

@misc{qwenmax2025,
  title={{Qwen3-Max: Just Scale it}},
  author={Qwen Team},
  year={2025},
  url={https://qwen.ai/blog?id=qwen3-max}
}

@misc{qwenembed2025,
  title={Qwen3-Embedding-{8B}},
  author={An Yang and Junyang Lin and Jingren Zhou and et al.},
  year={2025},
  url={https://huggingface.co/Qwen/Qwen3-Embedding-8B}
}

@misc{openai2024embedding3,
  title={{New embedding models and API updates}},
  author={OpenAI},
  year={2024},
  url={https://openai.com/index/new-embedding-models-and-api-updates/}
}

@misc{openai2026gpt51,
  title        = {{GPT‑5.1: A smarter, more conversational ChatGPT}},
  author       = {{OpenAI}},
  year         = {2026},
  url={https://openai.com/zh-Hans-CN/index/gpt-5-1/},
}

@inproceedings{sun2026exposing,
  title={Exposing the Cracks: Vulnerabilities of Retrieval-Augmented LLM-based Machine Translation},
  author={Sun, Yanming and Zhan, Runzhe and Cheang, Chi Seng and Wu, Han and Liu, Xuebo and Niu, Yuyao and Ye, Fengying and Lan, Kaixin and Chao, Lidia S and Wong, Derek F},
  booktitle={Proceedings of the AAAI Conference on Artificial Intelligence},
  volume={40},
  number={39},
  pages={33135--33143},
  year={2026}
}

@inproceedings{Ye-etal-2026-probing,
    title = "Probing Semantic Alignment, Lexical Invariance, and Syntactic Influence in LLM Metaphor Processing",
    author = "Fengying Ye and Shanshan Wang and Lidia S. Chao and Derek F. Wong",
    booktitle = "Proceedings of the 64th Annual Meeting of the Association for Computational Linguistics (Volume 1: Long Papers)",
    month = jul,
    year = "2026",
    address = "San Diego, California",
    publisher = "Association for Computational Linguistics",
}

\appendix
\section{Additional Language Pairs} \label{app:data_stats}

\begin{table*}[t]
\centering
\small
\setlength{\tabcolsep}{18pt}
\begin{tabular}{l r l r l r l r}
\toprule
\textbf{Pair} & \textbf{N} &
\textbf{Pair} & \textbf{N} &
\textbf{Pair} & \textbf{N} &
\textbf{Pair} & \textbf{N} \\
\midrule
Ar--De & 5  & Ar--Zh & 2  & Ja--Ko & 9  & Pt--Vi & 42 \\
Ar--En & 3  & De--Ko & 6  & Ja--Th & 68 & Th--Vi & 58 \\
Ar--Es & 3  & De--Th & 52 & Ja--Vi & 52 & Zh--Ko & 2  \\
Ar--Fi & 4  & De--Vi & 55 & Ko--Es & 2  & Zh--Th & 28 \\
Ar--Fr & 13 & En--Ko & 1  & Ko--Pl & 3  & Zh--Vi & 18 \\
Ar--Ja & 9  & En--Th & 28 & Ko--Pt & 9  &        &    \\
Ar--Ko & 13 & En--Vi & 25 & Ko--Th & 8  &        &    \\
Ar--Pl & 2  & Es--Th & 54 & Ko--Vi & 8  &        &    \\
Ar--Pt & 8  & Es--Vi & 38 & Pl--Th & 42 &        &    \\
Ar--Th & 5  & Fi--Ko & 5  & Pl--Vi & 39 &        &    \\
Ar--Vi & 11 & Fi--Th & 62 & Pt--Th & 60 &        &    \\
\bottomrule
\end{tabular}
\caption{Additional language-pair composition in G-IdiomAlign. \textbf{N} denotes the count of aligned idiom pairs. Each pair is reported once using a canonical ordering.}
\label{tab:additional_pairs}
\end{table*}
In addition to the core G-IdiomAlign benchmark reported in Table~\ref{tab:pair_stats}, we construct a supplementary set of additional language pairs to broaden cross-lingual coverage in Table~\ref{tab:additional_pairs}. This extension introduces four new languages: Arabic (Ar), Korean (Ko), Thai (Th), and Vietnamese (Vi), and matches them with both the original benchmark languages and one another using the same gloss-pivoted pipeline.

Overall, this supplementary set contains 42 language pairs and 1,014 bidirectional rank-1 aligned idiom pairs. As in the core benchmark, these pairs are obtained through gloss-based candidate retrieval, bidirectional agreement, and distribution-aware filtering. Because coverage remains limited after precision-oriented filtering, we do not include these pairs in the main evaluation; instead, we release them to support future research on broader cross-lingual idiom alignment.

\section{Wiktionary Idiom Harvesting and Gloss Cleaning}
\label{app:scrape}

This appendix describes our cross-lingual harvesting framework for extracting idiom entries and their definition-based glosses from Wiktionary.

\paragraph{License.}
The data used in this section are derived from Wiktionary, a collaboratively constructed resource available under the Creative Commons Attribution-ShareAlike 3.0 License (CC BY-SA 3.0).

\paragraph{Harvesting.}
For each language, we enumerate idiom entry pages by traversing the corresponding idiom category on English Wiktionary and collecting linked entry pages across all ``next page'' partitions.
We optionally apply conservative filters to remove obvious auxiliary pages introduced by category-page organization.

\paragraph{Gloss extraction.}
From each entry page, we extract an example-free definition string as the gloss representation.
The extraction prioritizes sense-definition text and excludes usage examples and other non-definitional material.

\paragraph{Precision-oriented screening.}
We enforce a strict single-sense criterion: an entry is retained only if extraction yields exactly one candidate gloss string.
We then apply lightweight normalization to remove leading labels and standardize whitespace.

\section{Distribution-Aware Filtering: Implementation Details}
\label{app:dist_filter}

This appendix provides an implementation-level specification of the distribution-aware filtering step described in Section \ref{sec:data_construction}. The filtering procedure operates on the set of MNN-confirmed idiom pairs for a given language pair, each associated with its rank-1 gloss similarity score.

\paragraph{Input.}
For each \emph{unordered} language pair, after enforcing mutual nearest neighbors (MNN), we obtain a set of candidate alignments
\(
\mathcal{P}=\{(x_i,y_i)\}_{i=1}^{n},
\)
where each retained pair is associated with a rank-1 similarity score
\(s_i \in \mathbf{R}\),
computed as cosine similarity between the corresponding English-gloss embeddings (as defined in Section \ref{sec:data_construction}).
The MNN criterion itself is always enforced bidirectionally.

\paragraph{Equal-width binning.}
Let \(s_{\min}=\min_i s_i\) and \(s_{\max}=\max_i s_i\).
We partition the interval \([s_{\min}, s_{\max}]\) into 10 equal-width bins using 11 bin edges:
\[
e_j = s_{\min} + j \cdot \frac{s_{\max}-s_{\min}}{10}, \quad j=0,1,\dots,10.
\]

\paragraph{Modal-bin cutoff.}
Let \(c_j\) denote the number of MNN-confirmed pairs whose rank-1 scores fall into bin \(j\).
We define the modal bin as
\[
b = \arg\max_{j \in \{1,\dots,10\}} c_j.
\]
When multiple bins tie for the maximum count, ties are resolved by selecting the first maximizer returned by the implementation.

\paragraph{Filtering rule.}
An MNN-confirmed pair \((x_i,y_i)\) is retained if and only if its rank-1 similarity score falls in the modal bin or any higher bin:
\[
(x_i,y_i)\ \text{is kept} \iff \text{bin}(s_i) \ge b.
\]
Equivalently, the lower edge of the modal bin acts as a language-pair-specific cutoff.
As in the main text, similarity scores are treated as a relative diagnostic signal within each language pair, rather than as an absolute criterion of semantic correctness.

\section{Alignment Quality Evaluation Details}
\label{app:alignment_quality}

This appendix provides additional details on the annotation protocol, LLM prompting strategy, and statistical estimation used in Section~\ref{sec:alignment_quality}.

\paragraph{Annotation protocol.}
All idiom pairs are evaluated using a 3-point semantic equivalence scale: 2 (fully equivalent), 1 (partially equivalent), and 0 (non-equivalent). A score of 2 indicates that two idioms express the same core meaning and can be reasonably substituted in similar contexts; a score of 1 indicates partial equivalence with differences in tone, intensity, or pragmatic usage; and a score of 0 indicates non-equivalence. Annotators are instructed to focus on semantic meaning rather than literal form. Both human annotators and LLM judges follow the same annotation instructions and scoring criteria.

\paragraph{LLM prompting strategy.}
We implement LLM-based evaluation by embedding the annotation rubric directly into structured prompts. Each prompt takes as input a pair of idioms and their English glosses and requires the model to output an equivalence label (0/1/2).

To improve robustness and reduce prompt sensitivity, we adopt a multi-view prompting strategy with complementary perspectives, including (i) direct semantic comparison, (ii) a substitutability-based test that evaluates whether the two idioms can be used interchangeably in similar contexts, and (iii) comparison of pragmatic function and strength. All prompts share the same core rules: prioritize English glosses over surface forms and avoid relying on literal similarity.

Prompts are shown below:

\begin{quote}
\small

\texttt{Task: }\\
\texttt{Given two expressions in different languages and their English glosses, assign one equivalence label:}\\
\texttt{- 2 = Correct Equivalent (same core meaning; substitutable in similar contexts)} \\
\texttt{- 1 = Partially Correct (overlapping meaning but differences in tone or usage)}\\
\texttt{- 0 = Incorrect (different core meaning or function).}

\texttt{- Do not rely on literal similarity.}

\texttt{Input:}\\
\texttt{- Idiom A: <IDIOM\_A>}\\
\texttt{- GLOSS A: <GLOSS\_A\_EN>}\\
\texttt{- Idiom B: <IDIOM\_B>}\\
\texttt{- GLOSS B: <GLOSS\_B\_EN>}\\

\texttt{Output:}\\
\texttt{- Equivalence: <0/1/2>}
\end{quote}

\begin{quote}
\small

\texttt{Task: }\\
\texttt{Judge equivalence using a substitutable test.}\\
\texttt{Step 1: Based on the English glosses, imagine 2 short English contexts where this meaning is used.}\\
\texttt{Step 2: Decide if A and B could reasonably substitute each other in those contexts.}\\
\texttt{- 2 = Correct Equivalent (same core meaning; substitutable in similar contexts)} \\
\texttt{- 1 = Partially Correct (overlapping meaning but differences in tone or usage)}\\
\texttt{- 0 = Incorrect (different core meaning or function).}

\texttt{- Do not rely on literal similarity.}

\texttt{Input:}\\
\texttt{- Idiom A: <IDIOM\_A>}\\
\texttt{- GLOSS A: <GLOSS\_A\_EN>}\\
\texttt{- Idiom B: <IDIOM\_B>}\\
\texttt{- GLOSS B: <GLOSS\_B\_EN>}\\

\texttt{Output:}\\
\texttt{- Equivalence: <0/1/2>}
\end{quote}

\begin{quote}
\small

\texttt{Task: }\\
\texttt{Judge equivalence by explicitly comparing pragmatic function and strength.}\\
\texttt{- 2 = Correct Equivalent (same core meaning; substitutable in similar contexts)} \\
\texttt{- 1 = Partially Correct (overlapping meaning but differences in tone or usage)}\\
\texttt{- 0 = Incorrect (different core meaning or function).}

\texttt{- Do not rely on literal similarity.}

\texttt{Input:}\\
\texttt{- Idiom A: <IDIOM\_A>}\\
\texttt{- GLOSS A: <GLOSS\_A\_EN>}\\
\texttt{- Idiom B: <IDIOM\_B>}\\
\texttt{- GLOSS B: <GLOSS\_B\_EN>}\\

\texttt{Output:}\\
\texttt{- Equivalence: <0/1/2>}
\end{quote}
For each idiom pair, we obtain independent judgments from three LLMs (GPT-5.1, Gemini-2.5-pro, and Claude-4.5-Haiku). The final label is determined via majority voting across models. When all three models disagree, we assign a score of 1 to avoid over-claiming full equivalence while retaining borderline cases instead of discarding them. This combination of multi-prompt design and multi-model aggregation improves the robustness and stability of LLM-based judgments.

\begin{table*}[t]
\centering
\begin{tabular}{lcccccccc}
\hline
 & mean & std & min & \(p_{10}\) & \(p_{25}\) & \(p_{50}\) & \(p_{75}\) & \(p_{90}\) \\
\hline
Original \(s^{\text{orig}}\)
& 0.67 & 0.13 & 0.37 & 0.52 & 0.57 & 0.65 & 0.75 & 0.86 \\
Qwen3 \(s^{\text{Qwen3}}\)
& 0.78 & 0.10 & 0.35 & 0.66 & 0.72 & 0.78 & 0.85 & 0.91 \\
Shift \(\Delta\)
& 0.11 & 0.08 & -0.45 & 0.01 & 0.06 & 0.11 & 0.17 & 0.21 \\
\hline
\end{tabular}
\caption{Global similarity statistics for the OpenAI text-embedding-3-large and Qwen3-Embedding-8B, and the per-instance difference \(\Delta_i = s_i^{\text{Qwen3}} - s_i^{\text{orig}}\) over \(N=18{,}785\) aligned pairs.}
\label{tab:qwen-global-stats}
\end{table*}

\begin{table*}[t]
\centering
\begin{tabular}{lccccccccc}
\hline
Threshold \(t\)
& 0.518 & 0.567 & 0.617 & 0.666 & 0.716 & 0.766 & 0.815 & 0.865 & 0.914 \\
\hline
Original count
& 16898 & 14381 & 11378 & 8499 & 6027 & 4125 & 2803 & 1686 & 945 \\
Qwen3 count
& 18721 & 18545 & 17942 & 16606 & 14108 & 10583 & 6934 & 3933 & 1895 \\
\hline
\end{tabular}
\caption{Coverage under fixed similarity thresholds: number of aligned pairs with \(sim \ge t\) under each encoder.}
\label{tab:qwen-threshold-coverage}
\end{table*}

\paragraph{Sampling and statistical estimation.}
For Zh--En, we randomly sample 200 idiom pairs and evaluate them using human annotators. For all other language pairs, we sample 50 idiom pairs per language pair and evaluate them using the LLM-based protocol described above.

For non-Zh--En language pairs, each language pair is treated as one observation. We compute the mean strict and lenient accuracy across language pairs and report 95\% confidence intervals computed as t-intervals over language pairs.

For Zh--En, statistics are computed over individual samples ($n=200$). The reported accuracies correspond to the proportion of samples satisfying each criterion, where strict accuracy counts only score-2 pairs and lenient accuracy counts score-1 and score-2 pairs.

\paragraph{Results summary.}
For non-Zh--En language pairs, the mean strict accuracy is 0.685 (95\% CI [0.645, 0.724]) and the mean lenient accuracy is 0.923 (95\% CI [0.907, 0.940]). These correspond to 68.5\% fully equivalent pairs (score = 2) and 92.3\% partially or fully equivalent pairs (score $\geq 1$).

For Zh--En human evaluation, the strict accuracy is 0.655 (95\% CI [0.589, 0.721]) and the lenient accuracy is 0.895 (95\% CI [0.852, 0.938]). Despite being slightly more conservative, human evaluation yields results consistent with LLM-based estimates.

Overall, both LLM-based and human evaluations provide converging evidence that G-IdiomAlign achieves high semantic alignment quality, supporting the effectiveness of the proposed mining and filtering pipeline.

\section{Model Dependence of Similarity Scores}
\label{sec:cross-model}

We assess the sensitivity of gloss-based similarity scores to encoders by recomputing scores for the same aligned idiom pairs using Qwen3-Embedding-8B, an alternative multilingual embedding model. We examine three aspects: global score calibration, threshold-based coverage, and cross-encoder consistency in relative score structure.

\subsection{Global Statistics and Calibration Shift}
\label{app:qwen-global}

For the same \(N=18{,}785\) aligned pairs, we recompute cosine similarities between source and target glosses using Qwen3-Embedding-8B and compare them with the construction-time scores obtained using OpenAI text-embedding-3-large. Both encoders use cosine similarity over $L_2$-normalized embeddings.

Table~\ref{tab:qwen-global-stats} summarizes the original scores \(s^{\text{orig}}\), the recomputed scores \(s^{\text{Qwen3}}\), and the per-instance difference \(\Delta_i = s_i^{\text{Qwen3}} - s_i^{\text{orig}}\). Qwen3 produces systematically higher absolute similarity values than the original encoder (mean \(0.67 \rightarrow 0.78\); median \(0.65 \rightarrow 0.78\)), with an average shift of \(\Delta\mu=0.11\). However, the shift is heterogeneous across instances (\(p_{10}=0.01\), \(p_{90}=0.21\)) and includes negative values, indicating that the difference is not reducible to a simple global offset or rescaling.

\subsection{Threshold Sensitivity Under Calibration Shift}
\label{app:qwen-threshold}

To illustrate the practical effect of calibration differences, we count the number of aligned pairs satisfying \(s \ge t\) under fixed absolute thresholds \(t\). We sweep nine thresholds uniformly over the intersection of the two encoders' 10th--90th percentile score intervals.

Table~\ref{tab:qwen-threshold-coverage} shows that the number of retained pairs differs substantially across encoders at the same threshold. This shows that absolute similarity thresholds are not directly comparable across embedding models in this setting, supporting our treatment of similarity as a relative diagnostic signal in the main text.

\subsection{Embedding Consistency}
\label{app:qwen-corr}

Despite the shift in absolute similarity values, the relative score structure is largely preserved across encoders. Across all \(N=18{,}785\) aligned pairs, the two score sets are strongly correlated (Pearson \(r=0.807\)) and show substantial rank agreement (Spearman \(\rho=0.784\)). Thus, encoder choice has a larger effect on absolute calibration and threshold-based coverage than on the comparative ordering of aligned pairs.

%Figure~\ref{fig:qwen-perfile-scatter} aggregates this comparison at the language-pair level by plotting the mean similarity per language pair (construction-time encoder vs.\ Qwen3).
%The approximately monotonic relationship suggests that language pairs with higher average similarity under the construction-time encoder also tend to have higher average similarity under Qwen3, while the consistent upward displacement reflects encoder-specific calibration.
%\begin{figure}[t]
    %\centering
    %\includegraphics[width=0.70\linewidth]%{plots/per_file_mean_scatter.pdf}
    %\caption{Language-pair level agreement between the openAI text-embedding-3-large and Qwen3-Embedding-8B. Each point is a language pair, plotted by mean similarity under the openAI encoder (x-axis) versus mean similarity under Qwen3 (y-axis).}
    %\label{fig:qwen-perfile-scatter}
%\end{figure}

\section{Task 1 Distractor Generation}
This appendix reports the unified prompt template used to generate the three typed distractors for Task~1 (Multiple-Choice Idiom Equivalence). For each instance, the canonical target idiom from G-IdiomAlign is used as the reference option, while Qwen-Max is used \emph{only} to generate the remaining three distractors under a single prompt: Literal Translation Trap (LT), Lexical Cue Trap (LC), and Contextual Association Trap (CA).
\subsection{Prompt}\label{app:task1_prompt}
\paragraph{Model and decoding.}
We generate distractors using \texttt{Qwen-Max} with greedy decoding (temperature $=0$, top-$p=1$).

\paragraph{Prompt template.}
The following prompt is used verbatim in our implementation, with placeholders instantiated per instance.

\begin{quote}\small
\texttt{You are an expert linguist specializing in cross-cultural idiom translation and test design. Your task is to create a multiple-choice question dataset to test whether an AI model truly understands idioms or just relies on literal translation.}

\vspace{0.5em}
\texttt{Input Data:}\\
\texttt{- Source Idiom: <SOURCE\_IDIOM>}\\
\texttt{- Source Meaning: <SOURCE\_MEANING>}\\
\texttt{- Source Language: <SOURCE\_LANGUAGE>}\\
\texttt{- Target Language: <TARGET\_LANGUAGE>}

\vspace{0.5em}
\texttt{Task:}\\
\texttt{Generate 3 options for a multiple-choice question.}\\
\texttt{1. Option (Literal Translation Trap): A direct, word-for-word translation of the source in <TARGET\_LANGUAGE>.}\\
\texttt{2. Option (Lexical Cue Trap): A real idiom in <TARGET\_LANGUAGE> that shares only part of a salient keyword from the literal translation but has a completely DIFFERENT meaning.}\\
\texttt{3. Option (Contextual Association Trap): A real idiom in <TARGET\_LANGUAGE> that has a related context but opposite meaning.}

\vspace{0.5em}
\texttt{Constraints:}\\
\texttt{- The 'Lexical Cue Trap' should clearly reflect a salient lexical cue from the literal translation.}\\
\texttt{Output strictly in JSON format.}
\end{quote}

\paragraph{Distractor intent.}
The prompt defines three distractor types for diagnosis by error category:
(i) \textbf{LT (Literal Translation Trap)} is a word-by-word rendering of the source idiom into the target language, designed to be surface-faithful but not meaning-equivalent;
(ii) \textbf{LC (Lexical Cue Trap)} is a target-language idiom that overlaps with a salient lexical cue from the literal translation while conveying a different meaning;
(iii) \textbf{CA (Contextual Association Trap)} is a target-language idiom that is contextually related yet semantically opposite to the intended meaning.

\paragraph{Instance assembly and shuffling.}
For each benchmark alignment pair $(x,y)$ in G-IdiomAlign, we take the canonical reference target idiom $y$ as the correct option and populate the remaining three options with the generated LT/LC/CA distractors. We then shuffle the option order per instance to reduce positional bias.

\subsection{Validity of LLM-Generated Distractors}
\label{app:distractor_validity}
We conduct a control in which the model is shown only the answer options, without the question stem. To avoid a trivial signal, we exclude the literal-translation hard negatives in this control, since they are intentionally designed to be non-idiomatic. The goal is to test whether the model can systematically prefer the gold idiomatic option based on surface properties alone.

Across three runs, the average selection rate for the gold option type is 0.3567, compared with a random baseline of approximately 0.3333. A chi-square test does not show a significant deviation from a uniform distribution ($\chi^2 = 2.94$, $p = 0.23$). This suggests that there is no strong evidence that the model can reliably identify the gold option from stylistic cues alone. To further reduce superficial shortcuts, we also shuffle option order and normalize option formatting.

\begin{table*}[t]
\centering
\small
\setlength{\tabcolsep}{2.5pt}
\begin{tabular}{lcccc|cccc|cccc}
\toprule
& \multicolumn{4}{c}{\textbf{Zh-target}} &
  \multicolumn{4}{c}{\textbf{En-target}} &
  \multicolumn{4}{c}{\textbf{Other-targets}} \\
\cmidrule(lr){2-5} \cmidrule(lr){6-9} \cmidrule(lr){10-13}
\textbf{Model} &
\textbf{Corr} & \textbf{LT} & \textbf{LC} & \textbf{CA} &
\textbf{Corr} & \textbf{LT} & \textbf{LC} & \textbf{CA} &
\textbf{Corr} & \textbf{LT} & \textbf{LC} & \textbf{CA} \\
\midrule
DeepSeek-V3.2 (NT) & 56.96 & 23.06 & 14.31 & 5.67 & 56.16 & 27.83 & 10.83 & 5.19 & 47.53 & 40.33 & 8.72 & 3.42 \\
DeepSeek-V3.2 (T)  & 69.18 & 17.96 & 10.37 & 2.49 & 65.38 & 24.49 & 7.13  & 3.00 & 54.09 & 37.77 & 6.53 & 1.60 \\
Gemini-2.5-Pro     & 67.57 & 17.74 & 10.17 & 4.52 & 64.95 & 24.81 & 6.84  & 3.40 & 62.91 & 28.54 & 6.02 & 2.52 \\
Claude-4.5-Haiku   & 57.30 & 24.06 & 14.40 & 4.23 & 53.31 & 27.78 & 13.92 & 4.99 & 43.44 & 46.38 & 7.77 & 2.41 \\
Ministral-8B-Instruct & 28.51 & 26.18 & 26.78 & 18.53 & 28.97 & 30.03 & 24.97 & 16.03 & 27.76 & 36.55 & 20.64 & 15.06 \\
Qwen3-8B (NT)      & 33.72 & 33.50 & 24.72 & 8.06 & 33.62 & 31.01 & 24.84 & 10.52 & 23.99 & 55.99 & 14.68 & 5.34 \\
Qwen3-8B (T)       & 41.89 & 38.81 & 16.01 & 3.29 & 39.10 & 41.53 & 14.78 & 4.59 & 31.39 & 52.52 & 12.80 & 3.29 \\
\bottomrule
\end{tabular}
\caption{Task~1 outcome proportions by target-language regime. Each row reports the \textbf{micro} fraction (\%) of instances that are \textit{Correct} or correspond to choosing one of the three typed distractors: Literal Translation Trap (LT), Lexical Cue Trap (LC), and Contextual Association Trap (CA). Values sum to 100\% within each model--regime block (up to rounding).}
\label{tab:task1_outcome_table}
\end{table*}

\begin{table*}[t]
\centering
\small
\setlength{\tabcolsep}{4pt}
\renewcommand{\arraystretch}{1.15}
\begin{tabular}{
    >{\centering\arraybackslash}m{1.9cm}
    >{\raggedright\arraybackslash}m{7cm}
    >{\centering\arraybackslash}m{2.8cm}
    >{\centering\arraybackslash}m{3.1cm}
}
\toprule
\textbf{Source Idiom} & \centering\textbf{Options} & \textbf{Gold Target Idiom} & \textbf{Model Prediction} \\
\midrule

\multirow{4}{*}{一丈差九尺}
& (A) wide of the mark (correct)
& \multirow{4}{=}{\centering (A) wide of the mark}
& \multirow{4}{=}{\centering (A) wide of the mark \ding{51}} \\
& (B) one zhang short by nine chi (literal trap) & & \\
& (C) measure twice, cut once (lexical cue trap) & & \\
& (D) hit the nail on the head (contextual association trap) & & \\

\addlinespace[3pt]
\midrule
\addlinespace[3pt]

\multirow{4}{*}{殺人不眨眼}
& (A) fish-blooded (correct)
& \multirow{4}{=}{\centering (A) fish-blooded}
& \multirow{4}{=}{\centering (B) kill without blinking an eye\\ (literal trap) \ding{55}} \\
& (B) kill without blinking an eye (literal trap) & & \\
& (C) bat an eyelash (lexical cue trap) & & \\
& (D) have a heart of gold (contextual association trap) & & \\

\addlinespace[3pt]
\midrule
\addlinespace[3pt]

\multirow{4}{*}{仆心仆肺}
& (A) bend over backwards (correct)
& \multirow{4}{=}{\centering (A) bend over backwards}
& \multirow{4}{=}{\centering (C) heart and soul\\ (lexical cue trap) \ding{55}} \\
& (B) servant heart, servant lungs (literal trap) & & \\
& (C) heart and soul (lexical cue trap) & & \\
& (D) pull your punches (contextual association trap) & & \\

\addlinespace[3pt]
\midrule
\addlinespace[3pt]

\multirow{4}{*}{善罷甘休}
& (A) fold like a cheap suit (correct)
& \multirow{4}{=}{\centering (A) fold like a cheap suit}
& \multirow{4}{=}{\centering (D) hold a grudge\\ (related but not equivalent) \ding{55}} \\
& (B) willingly stop and sweetly rest (literal trap) & & \\
& (C) sweet tooth (lexical cue trap) & & \\
& (D) hold a grudge (contextual association trap) & & \\

\bottomrule
\end{tabular}
\caption{Representative examples for Task~1 (multiple-choice). Each instance contains one correct target idiom and three typed distractors. \ding{51} indicates a correct prediction and \ding{55} an incorrect one.}
\label{tab:task1_case_study}
\end{table*}

\paragraph{Manual verification of distractor-type validity.}
We additionally manually inspect 200 questions to verify whether the generated distractors match their intended categories (e.g., LT, LC, and CA). We adopt a strict per-question criterion: a question is counted as valid only if all four options match their intended types. Under this criterion, 163 out of 200 questions are valid, corresponding to an accuracy of 81.5\%.

These results indicate that the generated distractors largely satisfy the intended hard-negative constraints and are not trivially distinguishable by superficial signals alone.

\section{Task 2 Generation Prompts and Output Constraints}
\label{app:task2_prompts}

This appendix reports the prompts used for Task~2 (Gloss-Contrastive Generation) under the two input conditions: \textit{With-gloss} (source idiom plus its English gloss) and \textit{No-gloss} (source idiom only).

\paragraph{Prompt template (With-gloss).}
In the \textit{With-gloss} condition, we provide the source idiom together with its English gloss as an explicit semantic pivot, and ask the model to generate a meaning-equivalent idiom in the target language:

\begin{quote}\small
\texttt{Output the <TARGET\_LANGUAGE> idiom corresponding to "<SOURCE\_IDIOM>" with meaning "<GLOSS>".  
Return only one idiom and do not include any explanation or additional text.}
\end{quote}

\paragraph{Prompt template (No-gloss).}
In the \textit{No-gloss} condition, we provide only the source idiom and ask the model to generate the corresponding idiom in the target language:

\begin{quote}\small
\texttt{Output the <TARGET\_LANGUAGE> idiom corresponding to "<SOURCE\_IDIOM>".  
Return only one idiom and do not include any explanation or additional text.}
\end{quote}

\begin{table*}[t]
\centering
\small
\begin{tabular}{lccccc}
\toprule
\textbf{Model} & \textbf{MeanSim} & \textbf{Acc@0.65} & \textbf{Acc@0.70} & \textbf{Acc@0.75} & \textbf{Acc@0.80} \\
\midrule
\multicolumn{6}{l}{\textbf{No-gloss}} \\
\midrule
DeepSeek-V3.2 (NT) & 68.30 & 58.02 & 41.99 & 28.63 & 19.90 \\
DeepSeek-V3.2 (T)  & 72.31 & 67.64 & 49.01 & 34.10 & 24.66 \\
Gemini-2.5-Pro     & 68.71 & 61.33 & 44.06 & 29.98 & 19.70 \\
Claude-4.5-Haiku   & 70.90 & 65.47 & 45.47 & 30.25 & 20.47 \\
Ministral-8B-Instruct & 68.30 & 59.86 & 37.97 & 21.96 & 12.96 \\
Qwen3-8B (NT)      & 67.54 & 57.22 & 36.29 & 20.78 & 11.59 \\
Qwen3-8B (T)       & 68.53 & 60.26 & 39.03 & 23.42 & 14.04 \\
\midrule
\multicolumn{6}{l}{\textbf{With-gloss} (absolute; \(\Delta\) vs.\ No-gloss)} \\
\midrule
DeepSeek-V3.2 (NT) & 70.94 & 65.57 (+7.55) & 48.35 (+6.36) & 33.68 (+5.05) & 23.57 (+3.67) \\
DeepSeek-V3.2 (T)  & 74.24 & 72.50 (+4.86) & 54.90 (+5.89) & 39.71 (+5.61) & 29.63 (+4.97) \\
Gemini-2.5-Pro     & 72.54 & 68.22 (+6.89) & 51.09 (+7.03) & 36.86 (+6.88) & 27.11 (+7.41) \\
Claude-4.5-Haiku   & 72.94 & 70.58 (+5.11) & 52.16 (+6.69) & 36.41 (+6.16) & 25.66 (+5.19) \\
Ministral-8B-Instruct & 71.69 & 69.78 (+9.92) & 50.00 (+12.03) & 32.61 (+10.65) & 21.22 (+8.26) \\
Qwen3-8B (NT)      & 70.69 & 67.41 (+10.19) & 47.09 (+10.80) & 30.53 (+9.75) & 18.80 (+7.21) \\
Qwen3-8B (T)       & 71.40 & 69.20 (+8.94) & 49.16 (+10.13) & 32.36 (+8.94) & 20.86 (+6.82) \\
\bottomrule
\end{tabular}
\caption{Task~2 threshold sensitivity in a single table with two panels. MeanSim is \(100\times sim\) macro-averaged across directions. Acc@\(t\) is macro-averaged accuracy (\%) at threshold \(t\). In the \textit{With-gloss} panel, each Acc@\(t\) cell reports the absolute accuracy followed by the per-threshold gain over \textit{No-gloss} in parentheses (percentage points).}
\label{tab:task2_thresholds}
\end{table*}

\section{Task 1 Outcomes}
\subsection{Breakdown by Distractor Type}
\label{app:task1_outcomes}

This appendix reports the numeric outcome proportions corresponding to Figure~\ref{fig:task1_outcome_by_regime}. For each model and target-language regime (Zh-target, En-target, and Other targets), we decompose outcomes into the \textit{Correct} selection and three typed distractor selections: \textit{Literal Translation Trap} (LT), \textit{Lexical Cue Trap} (LC), and \textit{Contextual Association Trap} (CA). All values in Table~\ref{tab:task1_outcome_table} are regime-level \textbf{micro} proportions (instance-weighted within each regime) computed over the Task~1 evaluation instances for that regime; within each model--regime block, the four percentages sum to 100\% (up to rounding). These numeric breakdowns support the error-type comparisons discussed in Section \ref{sec:task1_error_types}.

\subsection{Case Study: Task 1 (Multiple-choice)} \label{app:task1_casestudy}
Table~\ref{tab:task1_case_study} presents representative examples from Task~1 to illustrate how the multiple-choice design probes different types of distractors. Each instance includes one correct target idiom and three typed distractors: a Literal Translation Trap, a  Lexical Cue Trap, and a Contextual Association Trap.

The examples show that models can succeed when they recover the intended figurative meaning (e.g., 一丈差九尺 $\rightarrow$ wide of the mark), but models may select (i) a literal translation that closely mirrors the source form (e.g., 殺人不眨眼), (ii) an option triggered by salient lexical cues (e.g., 仆心仆肺), or (iii) an expression that is semantically related but not equivalent (e.g., 善罷甘休).

These cases highlight that Task~1 not only evaluates overall accuracy, but also reveals distinct and interpretable failure modes through the use of typed distractors.

\begin{table*}[t]
\centering
\small
\begin{tabular}{lllp{6cm}c}
\hline
\textbf{Source Idiom} & \textbf{Gold Target} & \textbf{Model} & \textbf{Output} & \textbf{Sim.} \\
\hline

\multirow{3}{*}{修身養性}
& \multirow{3}{*}{build character}
& DeepSeek-V3.2 (T) & turn over a new leaf \ding{55} & 0.6750 \\
& & Claude-4.5-Haiku & cultivate one's moral character \ding{51} & 0.5561 \\
& & Qwen3-8B (T) & practice restraint \ding{55} & 0.4700 \\

\hline

\multirow{3}{*}{倒錢落海}
& \multirow{3}{*}{fool away}
& DeepSeek-V3.2 (T) & throw money down the drain \ding{51} & 0.7733 \\
& & Claude-4.5-Haiku & pour money down the drain \ding{51} (paraphrase) & 0.7617 \\
& & Qwen3-8B (T) & pour money down the drain \ding{51} (paraphrase) & 0.7617 \\

\hline

\multirow{3}{*}{de plantilla}
& \multirow{3}{*}{一技之長}
& DeepSeek-V3.2 (T) & 千篇一律 \ding{55} & 0.5789 \\
& & Claude-4.5-Haiku & 專業人士 \ding{55} & 0.5027 \\
& & Qwen3-8B (T) & 行家 \ding{55} & 0.6918 \\

\hline

\multirow{3}{*}{ir al grano}
& \multirow{3}{*}{cut to the chase}
& DeepSeek-V3.2 (T) & get to the point \ding{51} (paraphrase) & 0.8811 \\
& & Claude-4.5-Haiku & get down to brass tacks \ding{51} (paraphrase) & 0.8509 \\
& & Qwen3-8B (T) & cut to the chase \ding{51} & 1.0000 \\

\hline
\end{tabular}
\caption{Representative examples for Task~2 (open-ended generation) with embedding-based similarity scores. \ding{51} indicates correct or acceptable paraphrase; \ding{55} indicates incorrect predictions.}
\label{tab:task2_case_study}
\end{table*}

\section{Task 2 Outcomes}\label{app:task2_outcomes}
\subsection{Threshold Sensitivity} \label{app:task2_thresholds}

\begin{table*}[t]
\centering
\small
\setlength{\tabcolsep}{10pt}
\renewcommand{\arraystretch}{1.1}
\begin{tabular}{lccc ccc ccc}
\toprule
& \multicolumn{3}{c}{\textbf{EM}} 
& \multicolumn{3}{c}{\textbf{BLEU}} 
& \multicolumn{3}{c}{\textbf{ChrF}} \\
\cmidrule(lr){2-4} \cmidrule(lr){5-7} \cmidrule(lr){8-10}
\textbf{Model} 
& with & w/o & $\Delta$ 
& with & w/o & $\Delta$ 
& with & w/o & $\Delta$ \\
\midrule
DeepSeek-V3.2 (NT) &  9.36 &  8.62 & 0.74 & 16.82 & 15.45 & 1.37 & 23.90 & 22.60 & 1.30 \\
DeepSeek-V3.2 (T)  & 11.61 &  9.24 & 2.37 & 21.96 & 18.54 & 3.43 & 27.54 & 24.43 & 3.11 \\
Gemini-2.5-Pro     & 10.85 &  9.49 & 1.36 & 19.77 & 17.59 & 2.18 & 26.81 & 24.39 & 2.42 \\
Claude-4.5-Haiku   &  9.06 &  6.74 & 2.31 & 19.17 & 15.75 & 3.42 & 24.62 & 21.60 & 3.02 \\
Ministral-8B-Instruct &  6.15 &  4.62 & 1.52 & 15.31 & 12.14 & 3.16 & 20.47 & 17.46 & 3.01 \\
Qwen3-8B (NT)      &  4.08 &  4.05 & 0.03 & 13.22 & 11.65 & 1.57 & 18.79 & 17.46 & 1.34 \\
Qwen3-8B (T)       &  4.76 &  2.90 & 1.86 & 14.24 & 10.92 & 3.32 & 19.68 & 16.78 & 2.90 \\
\bottomrule
\end{tabular}
\caption{Auxiliary surface-form metrics for Task~2.}
\label{tab:task2_surface_metrics}
\end{table*}

We report Task~2 results under additional similarity thresholds \(t\in\{0.65,0.70,0.75,0.80\}\) in Table~\ref{tab:task2_thresholds}. MeanSim is reported as \(100\times sim\) and macro-averaged over directions. Acc@\(t\) is macro-averaged accuracy (\%) at threshold \(t\). For the \textit{With-gloss} panel, we additionally report the per-threshold improvement over \textit{No-gloss} in parentheses (percentage points).

\subsection{Case Study: Task 2 (Open-ended Generation)} \label{app:task2_casestudy}

Table~\ref{tab:task2_case_study} presents representative examples from Task~2 to illustrate model behavior in open-ended idiom generation. Unlike Task~1, this setting does not constrain models to a fixed set of options, and predictions are evaluated based on semantic equivalence or acceptable paraphrases of the target idiom.

The examples show that models can produce correct idiomatic expressions (e.g., ir al grano $\rightarrow$ cut to the chase) or acceptable paraphrases (e.g., 倒錢落海 $\rightarrow$ pour money down the drain), but errors often reflect difficulties in capturing the precise figurative meaning. In particular, models may generate expressions that are overly general (e.g., 修身養性 $\rightarrow$ practice restraint), semantically shifted (e.g., turn over a new leaf), or unrelated to the intended meaning (e.g., de plantilla).

\section{Task 2 Supplementary Evaluation}
\label{app:task2_metric_validity}

\paragraph{Human calibration.}
To provide a point of calibration for the embedding-based metric, we conduct a small human evaluation on Task~2 outputs. We manually annotate \(N=86\) examples for idiom-to-idiom correctness, obtaining an overall correct ratio of \(0.407\). Using these annotations as ground truth, embedding similarity yields an AUC of \(0.77\), with a 95\% confidence interval of \([0.67, 0.86]\), computed by nonparametric bootstrap resampling over the 86 examples.

At a representative threshold \(t=0.75\), the proxy achieves precision \(0.875\) (\(TP=7\), \(FP=1\)) with coverage \(9.3\%\). These values indicate that the metric is most reliable for identifying a relatively high-confidence subset of semantically correct outputs, rather than for exhaustively capturing all acceptable generations.

\paragraph{Surface-form metrics.}
As a complement to semantic matching, we also report auxiliary surface-form metrics against the canonical gold reference in Table~\ref{tab:task2_surface_metrics}. Specifically, we compute Exact Match (EM), and additionally report BLEU and ChrF as reference-based overlap measures. These metrics provide a view of whether a model output matches an attested or canonical idiom form.

As expected in open-ended generation, surface-form metrics should be interpreted with caution: semantically valid idiomatic paraphrases may still receive low scores if they differ from the canonical reference string. For this reason, we treat EM, BLEU, and ChrF as supplementary indicators. Although absolute EM values are low, the directional trends remain informative.

\section{Human Annotation Scheme for Gloss-Contrastive Generation}
\label{app:annotation_scheme}

We categorize each idiom chosen from Gloss-Contrastive Generation into one of five mutually exclusive labels $\{0,1,2,3,4\}$. Labels are assigned following the decision procedure below.

\paragraph{Label definitions.}

\begin{itemize}
    \item \textbf{Label 0 (Invalid / generation failure).}
    The output is empty, gibberish, or otherwise not interpretable as
    meaningful text in the target language.

    \item \textbf{Label 1 (Correct idiomatic translation).}
    The output is a fluent and semantically correct translation that
    realizes an appropriate idiomatic expression in the target language,
    conveying the intended meaning of the source idiom.

    \item \textbf{Label 2 (Literal word-by-word translation).}
    The output is semantically linked to the idiom’s surface form and
    appears to translate the idiom compositionally (word-by-word or
    phrase-by-phrase), rather than conveying the intended idiomatic
    meaning.

    \item \textbf{Label 3 (Meaning paraphrase, non-idiomatic form).}
    The output correctly conveys the intended meaning of the source
    idiom, but does so using a non-idiomatic paraphrase (i.e., not an
    idiom or conventional idiomatic expression in the target language).

    \item \textbf{Label 4 (Incorrect meaning).}
    The output is meaningful text in the target language but fails to
    convey the intended meaning of the source idiom, including cases of
    mistranslation, wrong sense, or unrelated content.
\end{itemize}

\paragraph{Decision procedure.}
Labels are assigned in the following order to ensure mutual exclusivity:
\begin{enumerate}
    \item If the output is not interpretable as meaningful text in the
    target language, assign \textbf{Label 0}.
    \item Otherwise, if the output is a fluent and correct idiomatic
    translation that appropriately realizes the source idiom in the
    target language, assign \textbf{Label 1}.
    \item Otherwise, if the output translates the idiom literally based
    on its surface form without conveying the intended idiomatic meaning,
    assign \textbf{Label 2}.
    \item Otherwise, if the output correctly conveys the intended meaning
    but does not use an idiomatic expression in the target language,
    assign \textbf{Label 3}.
    \item All remaining meaningful but semantically incorrect outputs
    are assigned \textbf{Label 4}.
\end{enumerate}

\section{Attention Diagnostics: Salient Heads and Layer Aggregation}
\label{app:attn_salience_defs}

This appendix defines the salient head and layer sets used in Table~\ref{tab:attn_heads_layers}. All computations use post-softmax attention weights from Qwen3-8B  using
\texttt{TransformerLens} under greedy decoding. We consider three subsets $u$: \textit{All}, \textit{Correct-only}, and \textit{Wrong-only}, under two conditions $c$: \textit{With-gloss} and \textit{No-gloss}.

\paragraph{Per-sample head scores.}
Let \(A^{(\ell,h)}_{i,t \rightarrow k}\) denote the attention weight from generation position \(t\) to key position \(k\) at layer \(\ell\) and head \(h\) for sample \(i\). Let \(\mathcal{T}_i\) be the set of template-defined generation positions and \(K_i\) the tracked token span (e.g., idiom span or gloss span). The head score is
\[
S_i^{(\ell,h)}
=
\frac{1}{|\mathcal{T}_i|}
\sum_{t \in \mathcal{T}_i}
\sum_{k \in K_i}
A^{(\ell,h)}_{i,t \rightarrow k},
\]
or the corresponding ratio-based variant for contrastive analyses.

\paragraph{Salient heads.}
For each sample \(i\), we retain the top-\(k\) heads ranked by \(S_i^{(\ell,h)}\) (\(k{=}10\)). We then count how often each head appears across samples and define the salient head set \(\mathcal{H}_{\mathrm{sal}}^{(c,u)}\) as the \(m{=}50\) most frequent heads.

\paragraph{Salient layers.}
We aggregate layers from the per-sample top-\(k\) heads. For each layer \(\ell\), we sum the number of per-sample top-\(k\) entries belonging to \(\ell\) across all samples, and retain the top-\(n\) layers by this count as the salient layer set \(\mathcal{L}_{\mathrm{sal}}^{(c,u)}\) (\(n{=}10\)).

\paragraph{Cross-condition overlap.}
For each subset \(u\), we report the Jaccard similarity between \textit{With-gloss} and \textit{No-gloss} salient sets, for both heads and layers. Higher values indicate greater structural overlap across conditions. These are the values reported in Table~\ref{tab:attn_heads_layers}.

\section{Token-level Attention Aggregation Details}
\label{app:attn_token_defs}

This section details the implementation of the token-level attention
diagnostics used in Section \ref{sec:attention}.~All attention weights are extracted from Qwen3-8B hooks and correspond to post-softmax
decoder self-attention under greedy decoding.

\paragraph{Attention extraction.}
For each generation, we collect the self-attention tensor at each layer
and head, yielding attention weights
$A^{(\ell,h)}_{t \rightarrow k}$ from generation query position $t$
to key token position $k$.
Attention is defined over the full prompt token sequence, including
template tokens.

\paragraph{Generation positions.}
Let $\mathcal{T}$ denote the set of template-defined generation positions
used for analysis.
In practice, $\mathcal{T}$ corresponds to the decoding positions
associated with the content-bearing segment of the output.
No additional filtering based on token type (e.g., punctuation)
is applied beyond this template-based selection.

\paragraph{Token span identification.}
Token spans corresponding to the idiom and (when present) the gloss
are identified by mapping character offsets in the prompt text to
token index ranges using the model tokenizer.
These spans are defined with respect to the full prompt tokenization
and are not re-indexed to exclude template tokens.

\paragraph{Token-level attention mass.}
For each key token position $k$, we compute the aggregated attention mass
\[
\bar{A}(k)
=
\frac{1}{|\mathcal{T}|LH}
\sum_{\ell=1}^{L}
\sum_{h=1}^{H}
\sum_{t \in \mathcal{T}}
A^{(\ell,h)}_{t \rightarrow k}.
\]
This quantity represents the average post-softmax attention mass
assigned to token $k$ across layers, heads, and selected generation
positions.

Using the aggregated attention mass $\bar{A}(k)$, we define the
token-level diagnostics reported in Section \ref{sec:attention}.

\end{CJK}
\end{document}